\documentclass[10pt,a4paper,conference]{IEEEtran}

\usepackage{algorithm}
\usepackage[noend]{algpseudocode}
\usepackage{color,array}
\usepackage{xcolor}

\usepackage{times}
\usepackage{epsfig}
\usepackage{graphicx}
\usepackage{amsmath}
\usepackage{amssymb}
\usepackage[marginal]{footmisc}

\usepackage[font=normalsize,labelfont=bf]{caption}
\definecolor{bg}{rgb}{0.66, 0.66, 0.66}
\definecolor{vtl}{RGB}{189,198,255}

\definecolor{vrev}{RGB}{1,255,254}

\definecolor{vstrt}{RGB}{255,238,232}

\definecolor{talk}{RGB}{255,0,246}

\definecolor{opndr}{RGB}{107,104,130}

\usepackage{cite}

\ifCLASSINFOpdf
\else
\fi

\begin{document}

\title{Multi-feature Co-learning for Image Inpainting}

\author{\IEEEauthorblockN{Jiayu Lin$^\dagger$, Yuan-Gen Wang$^\dagger$\IEEEauthorrefmark{1}, Wenzhi Tang$^\ddagger$, and Aifeng Li$^\ddagger$}
\IEEEauthorblockA{$^\dagger$\textit{School of Computer Science and Cyber Engineering, Guangzhou University, Guangzhou, China}\\
$^\ddagger$\textit{Network Center, Guangzhou University, Guangzhou, China}\\
}}




\maketitle

\begin{abstract}
Image inpainting has achieved great advances by simultaneously leveraging image structure and texture features. However, due to lack of effective multi-feature fusion techniques, existing image inpainting methods still show limited improvement. In this paper, we design a deep multi-feature co-learning network for image inpainting, which includes Soft-gating Dual Feature Fusion (SDFF) and Bilateral Propagation Feature Aggregation (BPFA) modules. To be specific, we first use two branches to learn structure features and texture features separately. Then the proposed SDFF module integrates structure features into texture features, and meanwhile uses texture features as an auxiliary in generating structure features. Such a co-learning strategy makes the structure and texture features more consistent. Next, the proposed BPFA module enhances the connection from local feature to overall consistency by co-learning contextual attention, channel-wise information and feature space, which can further refine the generated structures and textures. Finally, extensive experiments are performed on benchmark datasets, including CelebA, Places2, and Paris StreetView. Experimental results demonstrate the superiority of the proposed method over the state-of-the-art. The source codes are available at https://github.com/GZHU-DVL/MFCL-Inpainting.\footnotetext{\IEEEauthorrefmark{1}Email address of the corresponding author: wangyg@gzhu.edu.cn. This paper has been accepted for presentation in ICPR 2022.}

\end{abstract}

\IEEEpeerreviewmaketitle

\section{Introduction}

Image inpainting \cite{bertalmio2000image} aims at reconstructing damaged regions or removing unwanted regions of images while improving the visual aesthetics of the inpainted images, which has been widely used in low-level vision tasks, such as restoring corrupted photos and object removal. The main challenge of image inpainting is how to generate reasonable structures and realistic textures. Traditional image inpainting, such as patch-based methods \cite{efros2001image, barnes2009patchmatch}, fill out the hole with the most similar patch as the to-be-inpainted patch by searching on undamaged region. Since the similarity is computed on pixel domain of the image, these methods fail to generate objects with strong semantic information.

Deep learning-based methods have shown remarkable performance on image inpainting tasks \cite{pathak2016context, nazeri2019edgeconnect, ren2019structureflow, xiong2019foreground}. They can generate semantically consistent results by understanding high-level features of the images. Among these methods, the encoder-decoder architecture has been widely developed since this architecture can extract semantic features and generate visually pleasing contents even if the hole is large. The variants of encoder-decoder architecture like U-Net  \cite{yan2018shift} are also developed for image inpainting, which can enhance the feature connection between the encoder and the decoder. For irregularly corrupted images, Liu \textit{et al.} \cite{liu2018image} and  Yu \textit{et al.} \cite{yu2019free} proposed to force the network to exploit valid pixels only, obtaining considerable performance. Nevertheless, the above methods do not take full use of the image structure features so that their generative networks are difficult to generate the reasonable structures. This results in blur and artifacts around the hole boundaries.


\begin{figure}[t!]
\begin{center}
\includegraphics[width=1\linewidth]{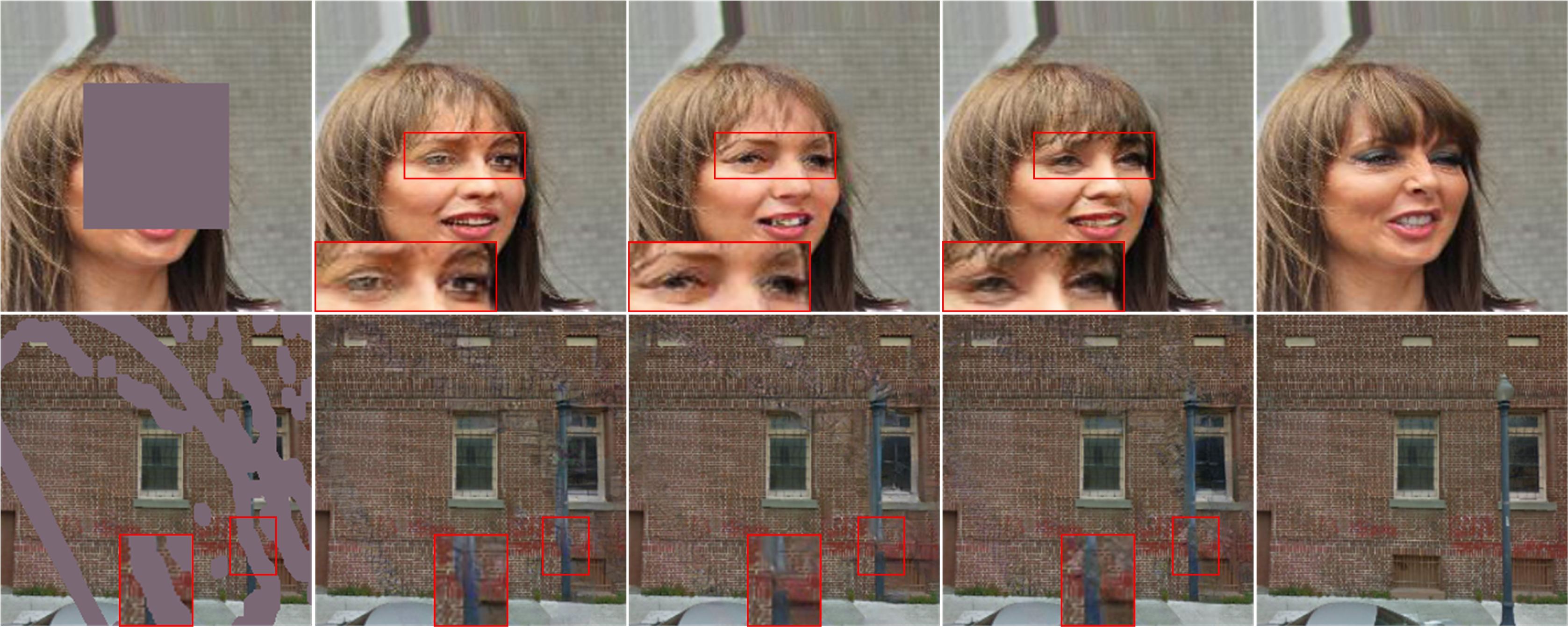}
\end{center}
\vspace{-3mm}
\caption{Visual results of the proposed method. From left to right: (a) Input images with holes. (b) Inpainted images without SDFF. (c) Inpainted images without BPFA. (d) Inpainted images with SDFF and BPFA. (e) Ground-truth images. We can observe that the inpainted images by our method have visually pleasing content.}
\label{fig:sample_frames}
\end{figure}

To overcome the above problems, several methods were proposed to use structural knowledge. For example, Nazeri \textit{et al.} \cite{nazeri2019edgeconnect} designed EdgeConnect with a two-stage model: edge generator and image completion generator. The edge generator is trained to predict the full edge map, while the image completion generator exploits the edge map as the structure prior to reconstruct the final image. Similar to EdgeConnect, Ren \textit{et al.} \cite{ren2019structureflow} designed  StructureFlow by using the edge-preserved smoothing technique \cite{xu2012structure}, making the structure reconstruction contain more information, such as image color. However, these methods do not simultaneously employ the structure and texture features, thereby leading to the inconsistent structures and textures of the output images. Based on this observation, Liu \textit{et al.} \cite{liu2020rethinking} designed MED to learn structures and textures separately by a mutual encoder-decoder, in which the deep-layer features are learned as structures, meanwhile the shallow-layer features are learned as textures. Although the consistency between structures and textures is improved, there have still been two major shortcomings: 1) The relationship between the structures and textures is not fully considered, resulting in limited consistency between them. 2) The context of an image is not sufficiently utilized, leading to  insufficient connection from local feature to overall consistency.

Motivated by these two shortcomings, we propose a multi-feature co-learning method for image inpainting. Specifically, we present two novel modules: 1) A Soft-gating Dual Feature Fusion (SDFF) module is designed to reorganize structure and texture features so that their consistency is greatly enhanced.  2) A Bilateral Propagation Feature Aggregation (BPFA) module is designed to capture the connection between contextual attention, channel-wise information, and feature space. This BPFA greatly enhances the connection from local feature to overall consistency. The major contributions of the proposed method are as follows:
\begin{itemize}
  \item We propose a novel SDFF module. With SDFF, the blur and artifacts around the holes are significantly reduced.
  \item We propose a novel BPFA module. With BPFA, the inpainted images show more rational structures and more detailed textures.
  \item Extensive experimental results demonstrate the superiority of the proposed method over the state-of-the-art.
\end{itemize}


\section{Related Works}
Traditional image inpainting methods are usually divided into two main categories: diffusion-based methods \cite{efros2001image, bertalmio2000image} and patch-based methods \cite{hays2007scene, darabi2012image}.  The first one propagates the appearance of adjacent content to fill out the missing regions. However, due to the limitation of search mechanism on adjacent content, there are obvious artifacts within images when facing large area masks. The second one fills in the missing region with the most similar patch as the to-be-inpainted patch. Although they can capture long-distance information, it is difficult to generate semantically reasonable images due to the lack of high-level structure understanding.

Deep learning-based methods \cite{yu2018generative, lahiri2020prior, zeng2020high, liu2021pd, peng2021generating, liao2021image} have been widely explored in the field of image inpainting. Pathak \emph{et al} \cite{pathak2016context} firstly developed encoder-decoder architecture and adversarial training for image inpainting. Iizuka \emph{et al} \cite{iizuka2017globally} overcame the information bottleneck defect by introducing a series of dilated convolution layers. Recently, Nazeri \emph{et al} \cite{nazeri2019edgeconnect} proposed EdgeConnect to generate possible edges and fill in the holes with precondition information. Like EdgeConnect, Xiong \emph{et al} \cite{xiong2019foreground} designed a similar model by adopting a contour generator as structure prior instead of the edge generator. Ren \emph{et al} \cite{ren2019structureflow} utilized the edge-preserving smoothing method to obtain sharp edges and low-frequency structures. Yang \emph{et al} \cite{yang2020learning} proposed a multi-task learning framework by introducing structure embedding to generate refined structures. Liu \emph{et al} \cite{liu2020rethinking} designed a mutual encoder-decoder network to learn the structure and texture features separately. Peng \emph{et al} \cite{peng2021generating} presented the conditional autoregressive network and structure attention module, which can learn the distribution of structure features and capture the distance relationship between structures, respectively. However, the above-reviewed methods do not fully consider the relationship between the structure and texture features, it is difficult to generate the images with reasonable structures and sophisticated textures.

\section{Proposed method}

\begin{figure*}[ht!]
\begin{center}
\includegraphics[width=1\linewidth]{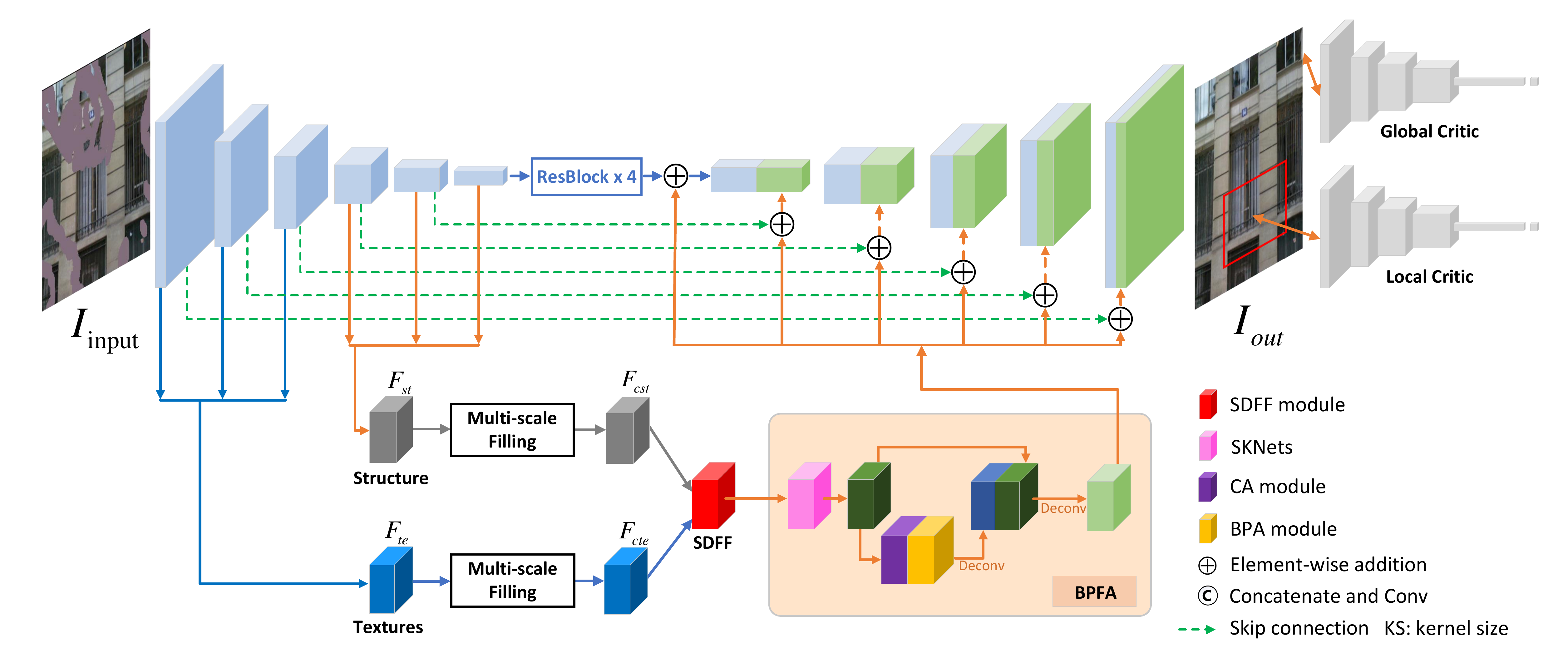}
\end{center}
\vspace{-6mm}
\caption{Description of the proposed pipeline. \textbf{Generator}: We propose a variant of U-Net architecture to jointing learn image structures and textures. The Soft-gating Dual Feature Fusion (SDFF) module and Bilateral Propagation Feature Aggregation (BPFA) module are designed to refine the generated structures and textures. \textbf{Discriminator}: We adopt the local and global discriminators to ensure from local to global content consistency.}
\label{fig:system}
\end{figure*}

The overall pipeline of the proposed method is shown in Fig. \ref{fig:system}, which is built upon the generative adversarial network. The generative network consists of mutual encoder-decoder, structure and texture branches, Soft-gating Dual Feature Fusion (SDFF), and Bilateral Propagation Feature Aggregation (BPFA). The discriminative network consists of global discriminator and local discriminator. By convention, the generative network aims to generate the inpainted images by co-learning image structures and textures, while the discriminative network aims to distinguish between the inpainted images and real ones. In the following, we describe the proposed network architecture and loss functions in detail.

\subsection{Generator} \label{tube_extraction}
The generator can be divided into the five parts: 1) The encoder consists of six convolutional layers. The three shallow-layer features are reorganized as texture features to represent image details. Meanwhile, the three deep-layer features are reorganized as structure features to represent image semantics. 2) We adopt two branches to separately learn the structure and texture features. 3) We design a SDFF module to fuse the structure and texture features generated by the above two branches. 4) We design a BPFA module to equalize the features between contextual attention, channel-wise information, and feature space. 5) The skip connection is used to supplement decoder features, which helps synthesize structure and texture branches to produce more sophisticated images.


\textbf{Structure and Texture Branches.}
The texture feature reorganized by shallow-layer convolution is denoted as $\boldsymbol{F}_{te}$ and the structure feature reorgnized by deep-layer convolution is denoted as $\boldsymbol{F}_{st}$. In each branch, three parallel streams are used to fill out the corrupted regions at multiple scales. For each stream, we replace all the vanilla convolutions with padding based partial convolution in order to better fill in irregular holes. Note that each stream consists of five convolutional layers, and the convolutional kernel sizes of the three streams are $3\times 3$, $5\times 5$ and $7\times 7$, respectively. We can obtain the filled features by first combining the output feature maps of the three streams and then mapping the combined features into the same size of the input feature. Here, we denote $\boldsymbol{F}_{cst}$ and $\boldsymbol{F}_{cte}$ as the outputs of the structure and texture branches, respectively. To ensure that the two branches focus on structures and textures respectively, we use two reconstruction losses, denoted as $\mathcal{L}_{\textrm{rst}}$ and $\mathcal{L}_{\textrm{rte}}$ respectively. The pixel-wise loss is defined as:
\begin{equation} \label{con:structureBranch}
\footnotesize{
 \mathcal{L}_{\textrm{rst}}=\left\|g\left(\boldsymbol{F}_{cst}\right)-\boldsymbol{I}_{st}\right\|_1,
}
\end{equation}
\begin{equation} \label{con:textureBranch}
\footnotesize{
 \mathcal{L}_{\textrm{rte}}=\left\|g\left(\boldsymbol{F}_{cte}\right)-\boldsymbol{I}_{gt}\right\|_1,
}
\end{equation}
where $g(\cdot)$ is the convolution operation with the kernel size of 1, which aims to map $\boldsymbol{F}_{cst}$ and $\boldsymbol{F}_{cte}$ to two color images respectively. $\boldsymbol{I}_{gt}$ and $\boldsymbol{I}_{st}$ denote the ground-truth image and its structure image, respectively. We follow \cite{ren2019structureflow} by using an edge-preserving smoothing method \cite{xu2012structure} to generate $\boldsymbol{I}_{st}$.

\begin{figure}
	\centering
	\includegraphics[width=1\linewidth, trim={0.5cm, 0.5cm, 0.5cm, 0.5cm}]{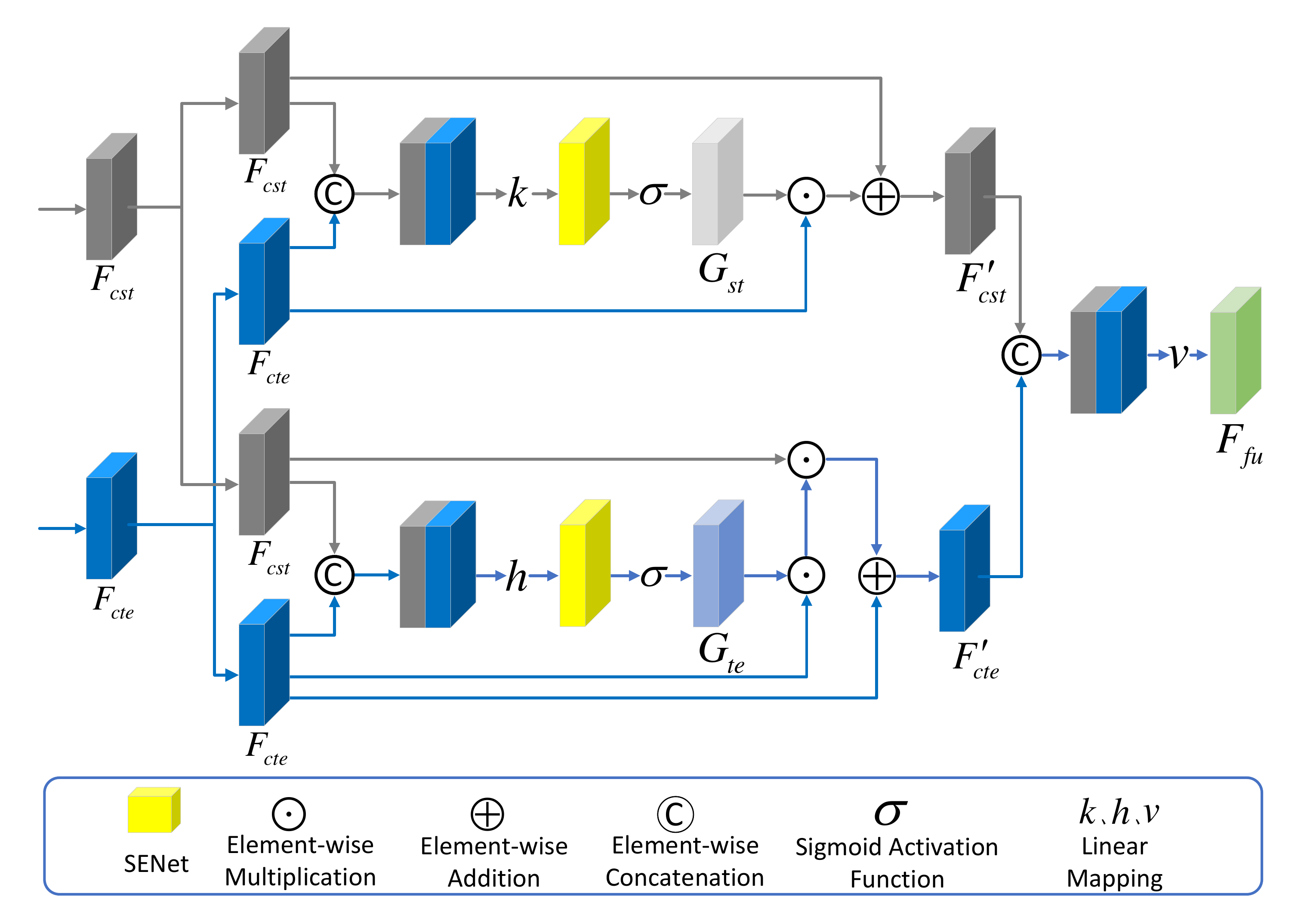}
	\vspace{-4mm}
	\caption{Description of Soft-gating Dual Feature Fusion, which can effectively fuse the structure and texture features.}
	\label{fig:SDFF module}
\end{figure}

\textbf{Soft-gating Dual Feature Fusion (SDFF).}
This module is designed to better exchange the structure features $\boldsymbol{F}_{cst}$ and texture features $\boldsymbol{F}_{cte}$ generated by the above two branches, respectively. The exchange is implemented by utilizing a soft gating way to dynamically adjust the fusion ratio between the structure and texture features. Fig. \ref{fig:SDFF module} illustrates the proposed SDFF module. Specifically, in order to construct the structure-guided texture features, we utilize the soft gating $\boldsymbol{G}_{te}$ to control a degree of refining the texture information. The soft gating can be defined as:
\begin{equation}
\footnotesize{
 \boldsymbol{G}_{te}=\sigma\left(SE\left(h\left(\left[\boldsymbol{F}_{cst}, \boldsymbol{F}_{cte}\right]\right)\right)\right),
}
\end{equation}
where $h(\cdot)$ is a convolution layer with the kernel size of 3, $SE(\cdot)$ is a squeeze and excitation operation \cite{hu2018squeeze} to capture important channel information, and $\sigma(\cdot)$ is the Sigmoid activation function. With soft gating $\boldsymbol{G}_{te}$, we can dynamically merge $\boldsymbol{F}_{cst}$ into $\boldsymbol{F}_{cte}$ by
\begin{equation}
\footnotesize{
 \boldsymbol{F^{\prime}}_{cte}=\alpha\left(\beta\left(\boldsymbol{G}_{te}\odot\boldsymbol{F}_{cte}\right)\odot\boldsymbol{F}_{cst}\right)\oplus\boldsymbol{F}_{cte},
}
\end{equation}
where $\alpha$ and $\beta$ are two learnable parameters, $\odot$ and $\oplus$ denote element-wise multiplication and addition respectively.

Similarly, the texture-guided structure features $\boldsymbol{F^{\prime}}_{cst}$ can be calculated as follows:

\begin{figure*}[ht!]
\begin{center}
\includegraphics[width=1\linewidth]{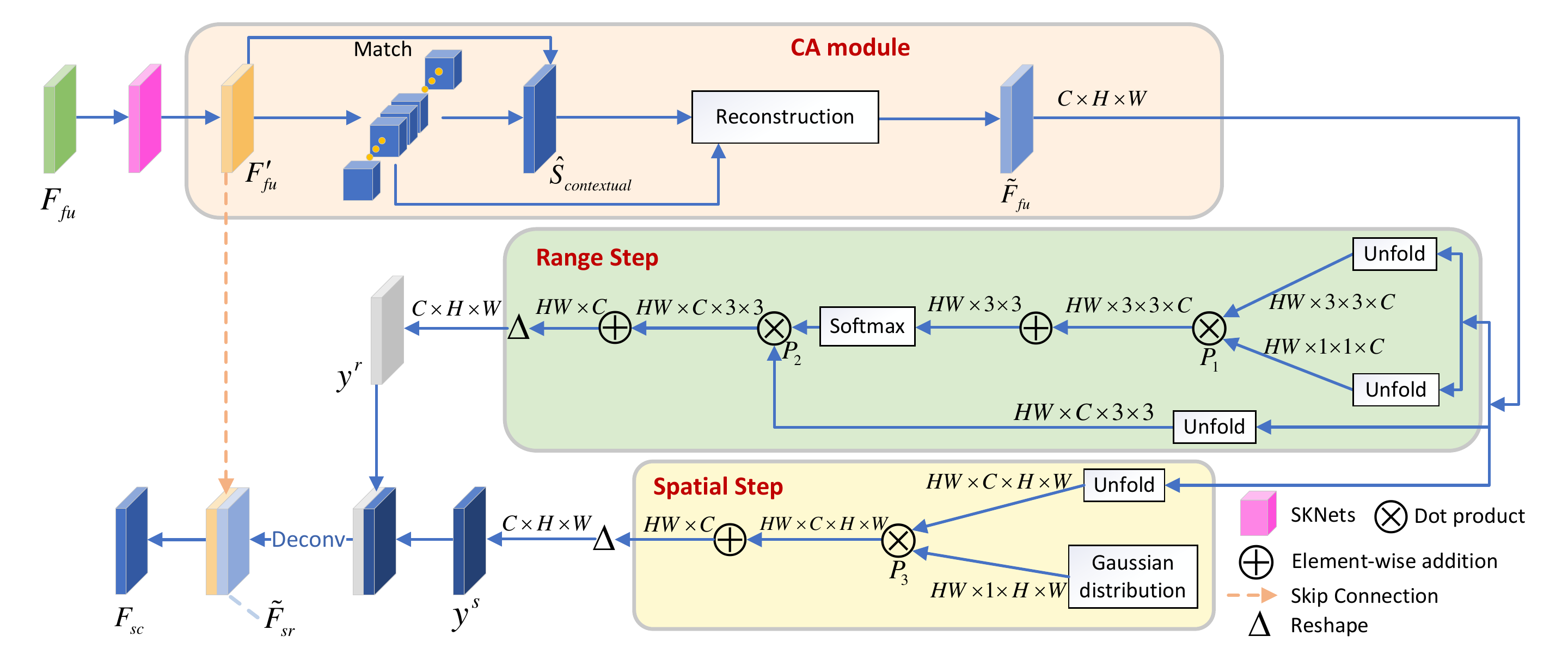}
\end{center}
\vspace{-4mm}
\caption{Description of the Bilateral Propagation Feature Aggregation (BPFA) module. It can enhance the connection from local feature to overall consistency.}
\label{fig:BPFA module}
\end{figure*}

\begin{equation}
\footnotesize{
 \boldsymbol{G}_{st}=\sigma\left(SE\left(k\left(\left[\boldsymbol{F}_{cst}, \boldsymbol{F}_{cte}\right]\right)\right)\right),
}
\end{equation}
\begin{equation}
\footnotesize{
 \boldsymbol{F^{\prime}}_{cst}=\gamma\left(\boldsymbol{G}_{st}\odot\boldsymbol{F}_{cte}\right)\oplus\boldsymbol{F}_{cst},
}
\end{equation}
where $k(\cdot)$ is a convolution layer with the kernel size of 3 and $\gamma$ is a learnable parameter. Thus, we can concatenate $\boldsymbol{F^{\prime}}_{cte}$ and $\boldsymbol{F^{\prime}}_{cst}$, and use a convolution layer $v$ with the kernel size of 1 to generate the integrated feature map $\boldsymbol{F}_{fu}$:
\begin{equation}
\footnotesize{
 \boldsymbol{F}_{fu}=v\left(\left[\boldsymbol{F^{\prime}}_{cst}, \boldsymbol{F^{\prime}}_{cte}\right]\right).
}
\end{equation}

\textbf{Bilateral Propagation Feature Aggregation (BPFA).}
This module is designed to co-learn contextual attention, channel-wise information, and feature space so as to enhance the overall consistency. Fig. \ref{fig:BPFA module} illustrates the proposed BPFA module in detail. Specifically, to capture channel-wise information, we use the Selective Kernel Convolution module of SKNet \cite{li2019selective} to generate the feature map $\boldsymbol{F^{\prime}}_{fu}$. Inspired by \cite{yu2018generative}, we introduce the Contextual Attention (CA) module to capture the correlation between feature patches. For a given input feature $\boldsymbol{F^{\prime}}_{fu}$, we divide it into non-overlapping patches with size 3$\times$3 and calculate the cosine similarity between these patches as:
\begin{equation}
\footnotesize{
 \boldsymbol{S}^{i,j}_{contextual}=\left \langle \frac{\boldsymbol{p}_{i}}{\left\|\boldsymbol{p}_{i}\right\|_{2}}, \frac{\boldsymbol{p}_{j}}{\left\|\boldsymbol{p}_{j}\right\|_{2}} \right \rangle,
}
\end{equation}
where $\boldsymbol{p}_{i}$ and $\boldsymbol{p}_{j}$ are the $i$-th and $j$-th patches of the input feature $\boldsymbol{F^{\prime}}_{fu}$, respectively. We utilize the Softmax function to get the attention score of each pair of patches:
\begin{equation}
\footnotesize{
 \boldsymbol{\hat{S}}^{i,j}_{contextual}=\mathrm{exp}\left(\frac{\boldsymbol{S}^{i,j}_{contextual}}{\sum_{j=1}^N\mathrm{exp}\left(\boldsymbol{S}^{i,j}_{contextual}\right)}\right),
}
\end{equation}
where $N$ is the total number of patches of the input feature $\boldsymbol{F^{\prime}}_{fu}$. Next, the attention score is used to compute  the feature patches $\widetilde{\boldsymbol{p}}_{i}, i=1,...,N$ by
\begin{equation}
\footnotesize{
 \widetilde{\boldsymbol{p}}_{i}=\sum_{j=1}^N\boldsymbol{p}_{j} \cdot \boldsymbol{\hat{S}}^{i,j}_{contextual}
}.
\end{equation}
The reconstructed feature map $\widetilde{\boldsymbol{F}}_{fu}$ can be obtained by directly reorganize all the feature patches.


In the range and spatial domains, we introduce the Bilateral Propagation Activation (BPA) module to generate the feature maps based on the range and spatial distances. The feature map of the range domain can be calculated as:

\begin{equation}
\footnotesize{
 \mathbf{y}^{r}_{i}=\frac{1}{C(x)}\sum_{j\in{v}}f\left(\mathbf{x}_{i},\mathbf{x}_{j}\right)\mathbf{x}_{j}, \label{con:range-domain}
}
\end{equation}
\begin{equation}
\footnotesize{
 f\left(\mathbf{x}_{i},\mathbf{x}_{j}\right)=(\mathbf{x}_{i})^{T}(\mathbf{x}_{j}), \label{con:dot-product}
}
\end{equation}
where we use the unfold function of PyTorch to reshape $\widetilde{\boldsymbol{F}}_{fu}$ to two kinds of vectors, which are of $HW \times 3 \times 3 \times C$ and $HW \times 1 \times 1 \times C$ dimensions. Here, $\mathbf{x}_{i}$ is the $i$-th output channel of the input feature $\widetilde{\boldsymbol{F}}_{fu}$, $\mathbf{x}_{j}$ is a neighboring channel at position $j$ around position $i$. The pairwise function $f(\cdot)$ is dot-product similarity. The operations from the unfold function to $P_{1}$ represent Eq. (\ref{con:dot-product}). $v$ is a neighboring region of position $i$ and its size is set to $3\times3$. For a given $i$, $\frac{1}{C(x)}f\left(\mathbf{x}_{i},\mathbf{x}_{j}\right)$ can be seen as the Softmax computation along dimension $j$. $C(x)$ is the number of channels in $\widetilde{\boldsymbol{F}}_{fu}$. Similarly, Eq. (\ref{con:range-domain}) represent the operations from on $\widetilde{\boldsymbol{F}}_{fu}$ until the position $P_{2}$. The feature map of the spatial domain can be calculated as:

\begin{equation}
\footnotesize{
 \mathbf{y}^{s}_{i}=\frac{1}{C(x)}\sum_{j\in{s}}g_{\alpha_{s}}\left(\left\|j-i\right\|_{1}\right)\mathbf{x}_{j}, \label{con:spatial-domain}
}
\end{equation}
where we use the unfold function of PyTorch to reshape $\widetilde{\boldsymbol{F}}_{fu}$ to a vector, which is of $HW \times C \times H \times W$ dimension. $g_{\alpha_{s}}$ is a Gaussian function to adjust the spatial contributions from neighboring region. We explore $j$ in a neighboring region $s$ for global propagation. In the experiment, $s$ is set to the same size as the input feature. Eq. \ref{con:spatial-domain} represent the operations from the unfold function to $P_{3}$. Therefore, we can obtain the feature maps $\mathbf{y}^{s}_{i}$ and $\mathbf{y}^{r}_{i}$ by the spatial and range similarity measurement methods, respectively. We can see that the bilateral propagation considers both local consistency via $\mathbf{y}^{s}_{i}$ and global consistency via $\mathbf{y}^{r}_{i}$. Each feature channel can be computed by
\begin{equation}
\footnotesize{
 \mathbf{y}_{i}=q\left(\left[\mathbf{y}^{s}_{i},\mathbf{y}^{r}_{i}\right]\right),}
\end{equation}
where $q$ denotes a convolution layer with the kernel size of 1. Next, the reconstructed feature map $\widetilde{\boldsymbol{F}}_{sr}$ is obtained by aggregating all feature channels $\mathbf{y}_{i}$ ($i=1,...,C(x)$). Finally, $\boldsymbol{F^{\prime}}_{fu}$ and $\widetilde{\boldsymbol{F}}_{sr}$ are concatenated and mapped to $\boldsymbol{F}_{sc}$ by
\begin{equation}
\footnotesize{
 \boldsymbol{F}_{sc}=z\left(\left[\boldsymbol{F^{\prime}}_{fu},\widetilde{\boldsymbol{F}}_{sr}\right]\right),
}
\end{equation}
where $z$ is a convolution layer with the kernel size of 1.

\def\swtwo{0.124\linewidth}
\begin{figure*}[!t]
\renewcommand{\tabcolsep}{0.5pt}
\centering
\begin{tabular}{cccccccc}
\includegraphics[width=\swtwo]{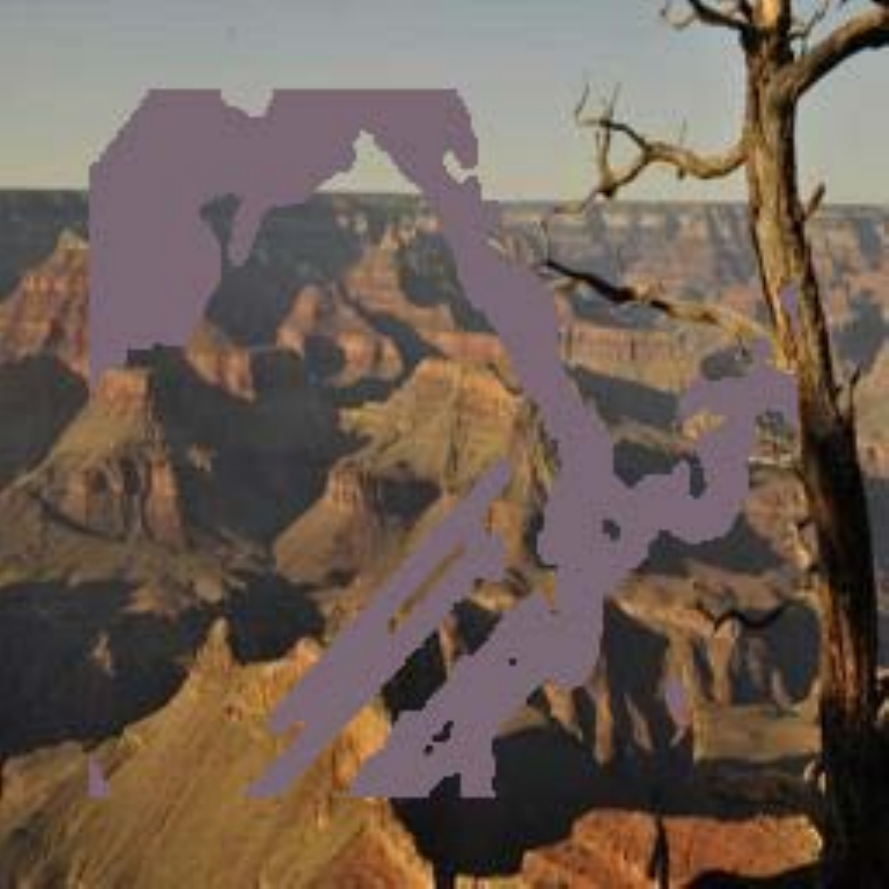}&
\includegraphics[width=\swtwo]{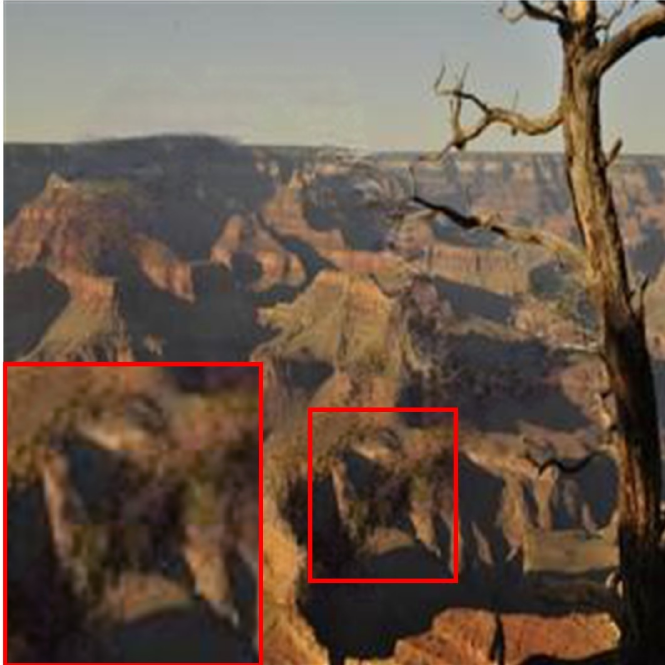}&
\includegraphics[width=\swtwo]{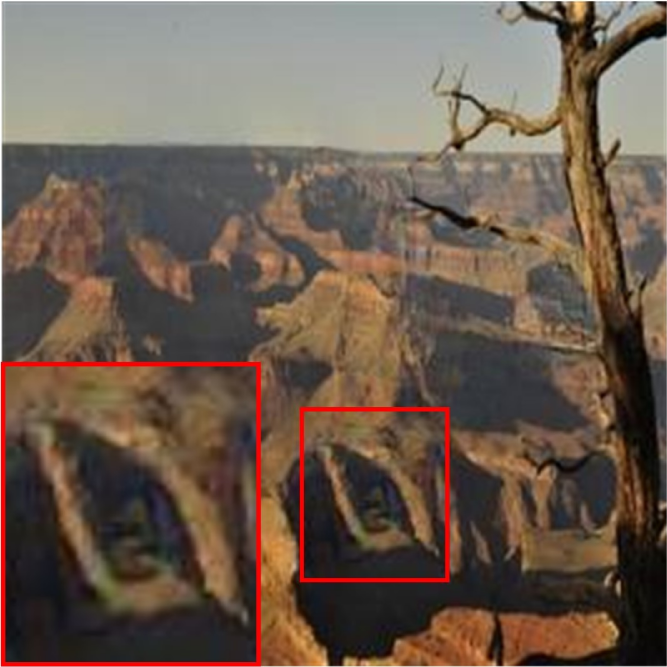}&
\includegraphics[width=\swtwo]{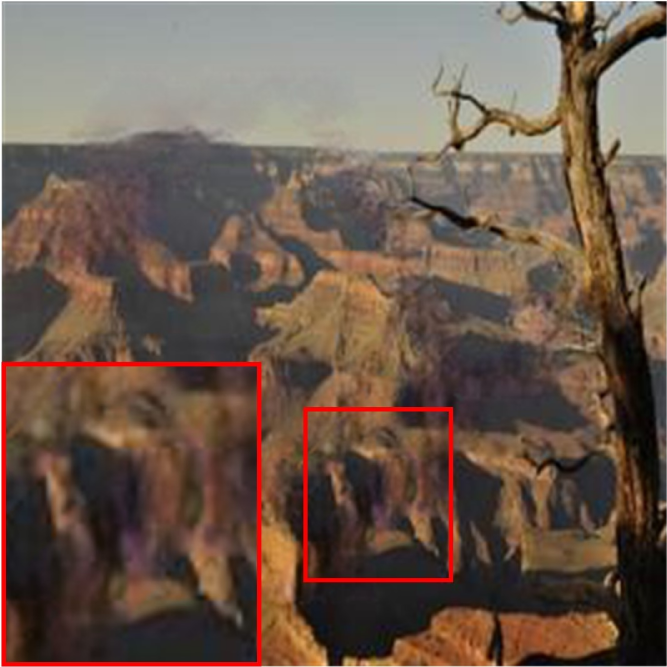}&
\includegraphics[width=\swtwo]{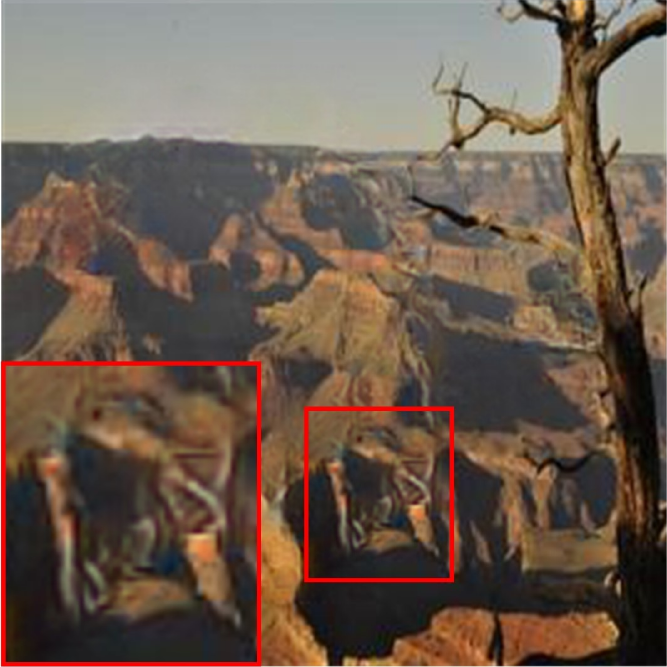}&
\includegraphics[width=\swtwo]{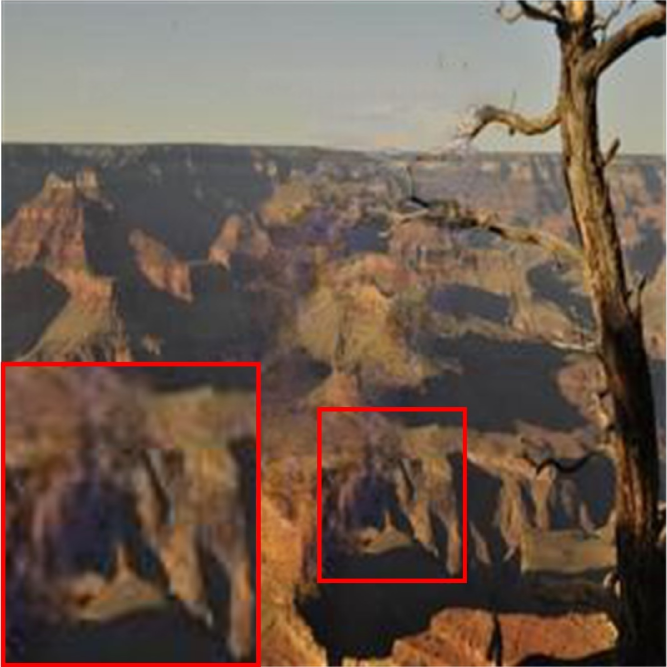}&
\includegraphics[width=\swtwo]{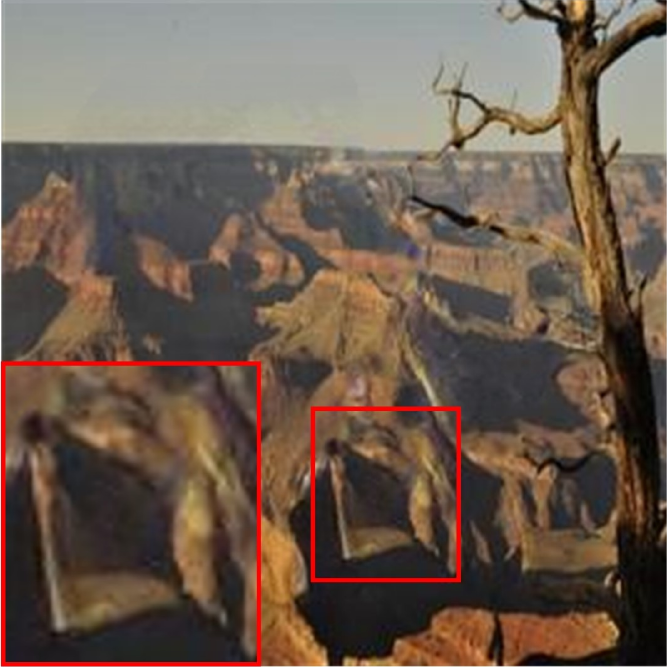}&
\includegraphics[width=\swtwo]{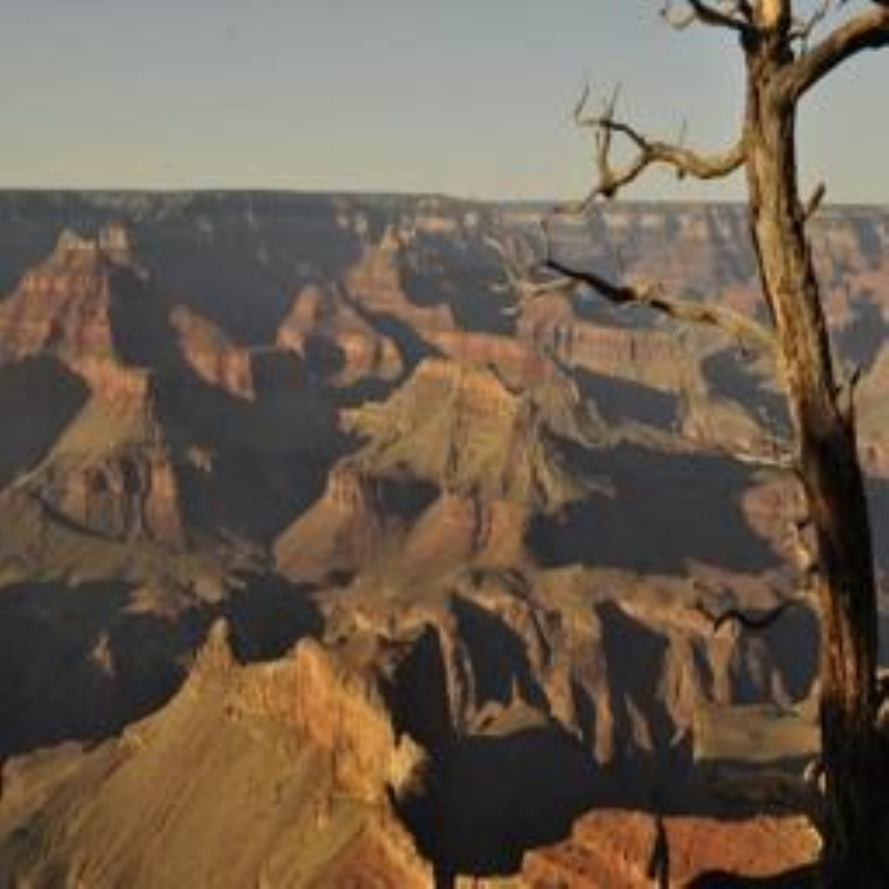}\\
\includegraphics[width=\swtwo]{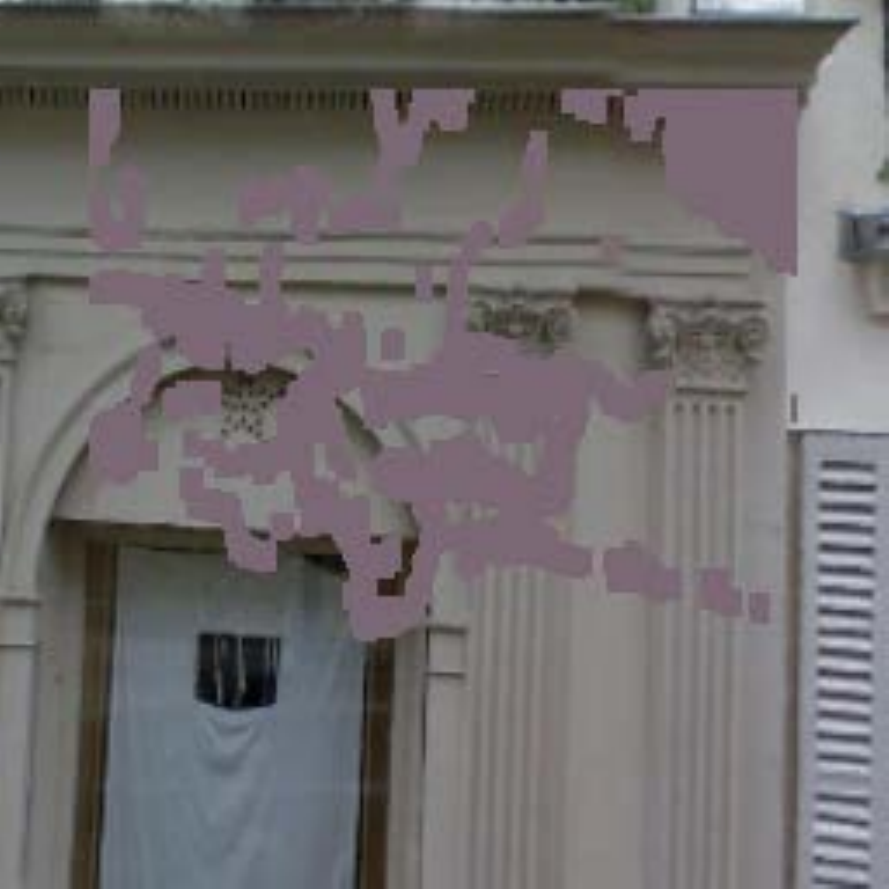}&
\includegraphics[width=\swtwo]{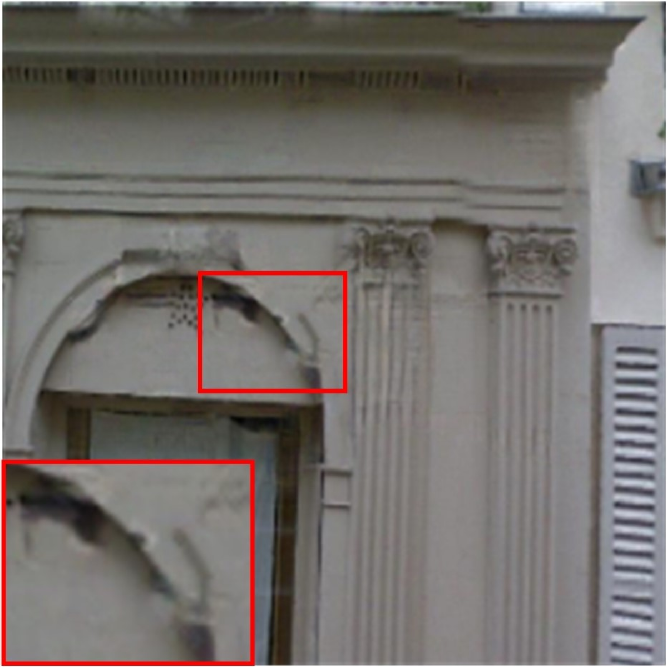}&
\includegraphics[width=\swtwo]{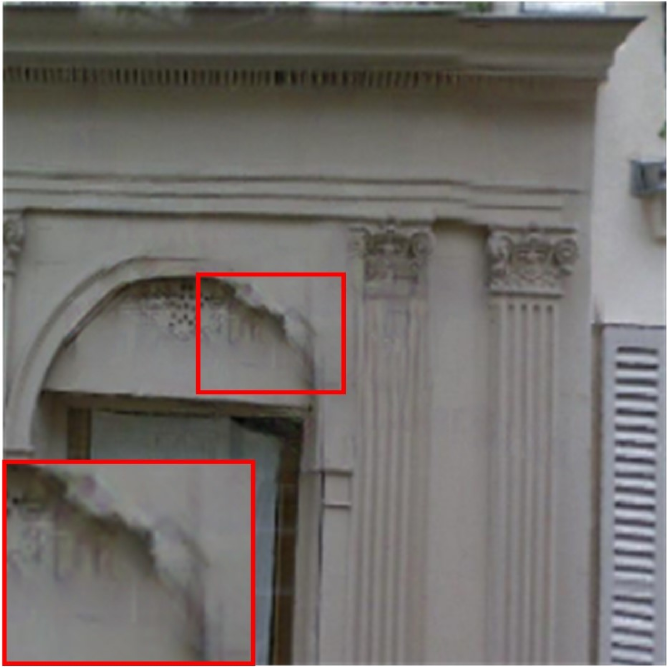}&
\includegraphics[width=\swtwo]{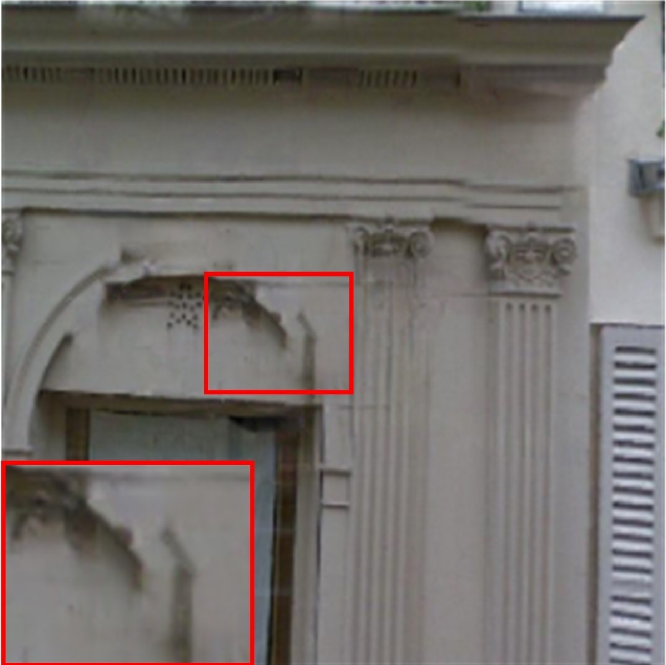}&
\includegraphics[width=\swtwo]{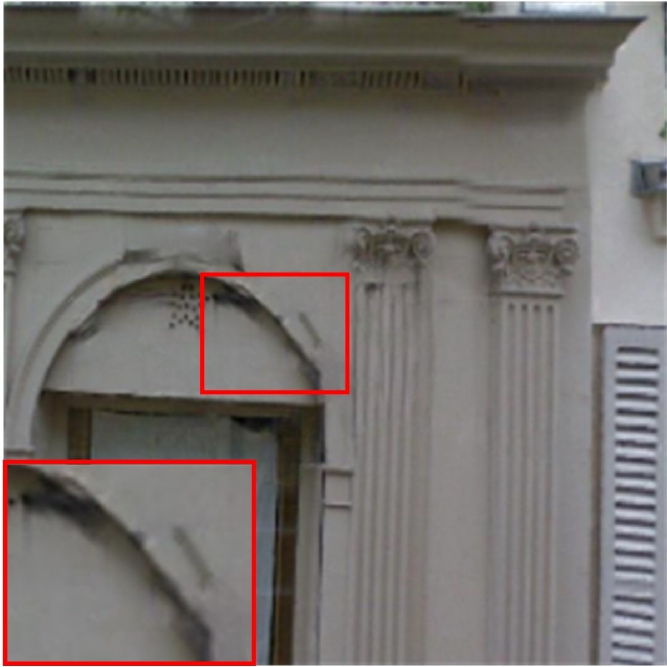}&
\includegraphics[width=\swtwo]{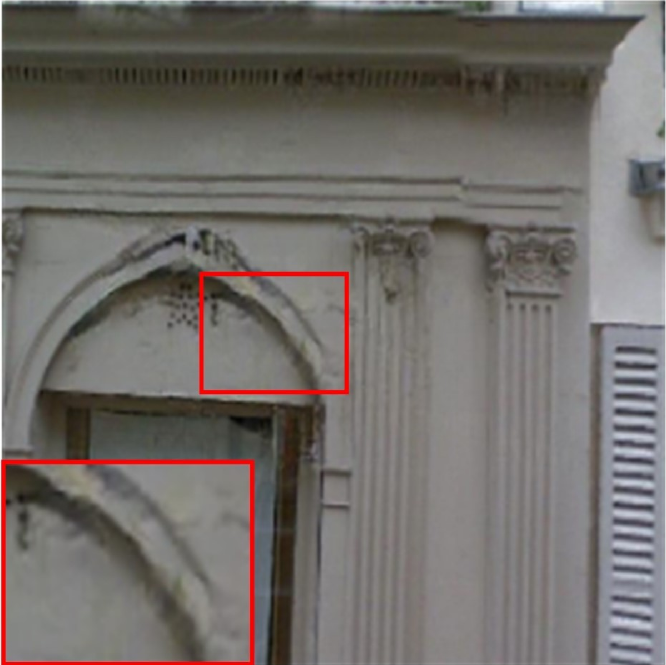}&
\includegraphics[width=\swtwo]{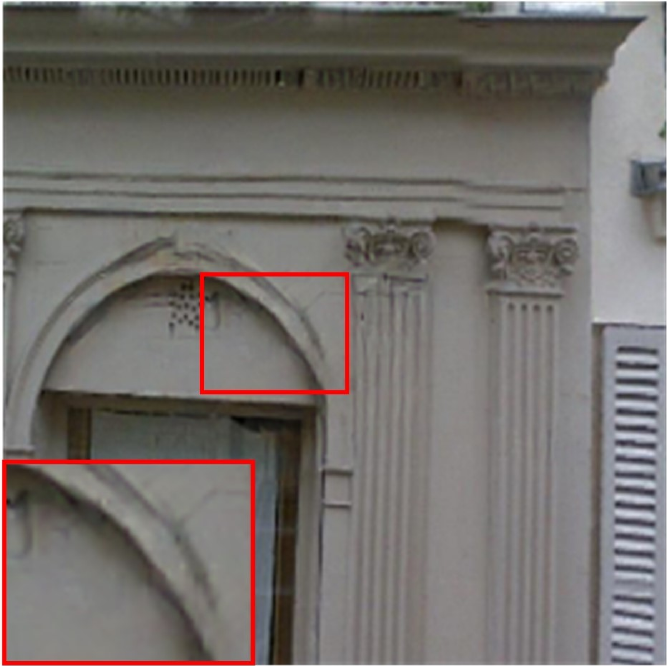}&
\includegraphics[width=\swtwo]{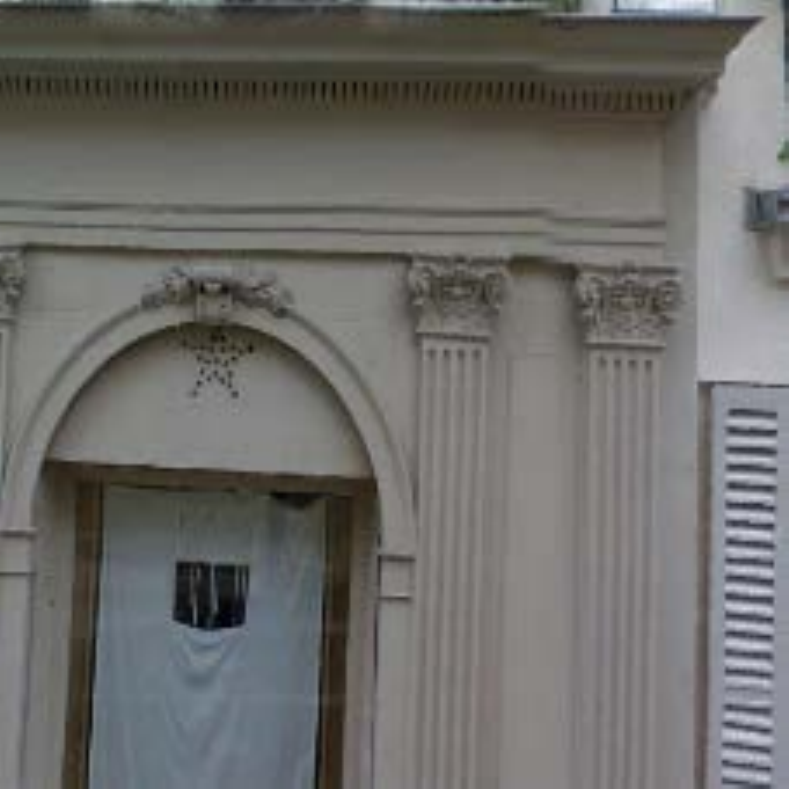}\\
(a) Input &(b) GC~\cite{yu2019free}&(c) EC~\cite{nazeri2019edgeconnect}&(d) MED~\cite{liu2020rethinking}&(e) RFR~\cite{li2020recurrent}&(f) DSI~\cite{peng2021generating}&(g) Ours&(h) GT
\end{tabular}
\vspace{-2mm}
\caption{Visual comparison on the Places2 (the first row) and  Paris StreetView (the second row) datasets.}
\label{imgirre}
\end{figure*}

\def\swtwo{0.124\linewidth}
\begin{figure*}[!t]
\renewcommand{\tabcolsep}{0.5pt}
\centering
\begin{tabular}{ccccccc}
\includegraphics[width=\swtwo]{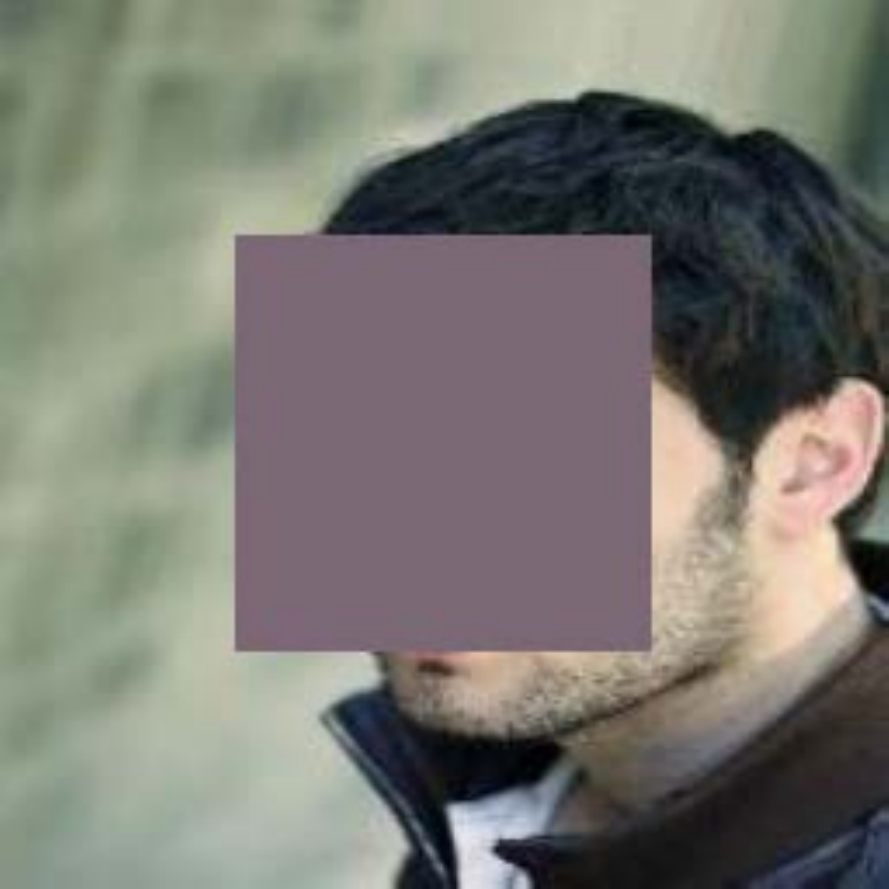}&
\includegraphics[width=\swtwo]{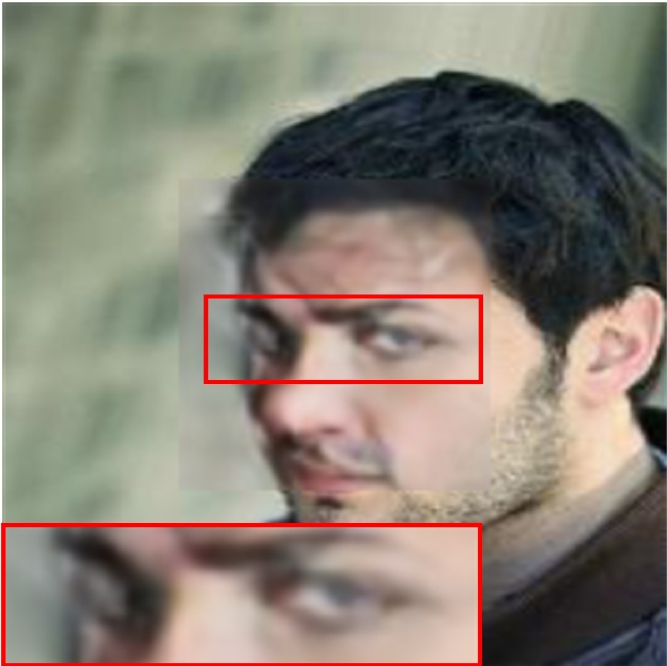}&
\includegraphics[width=\swtwo]{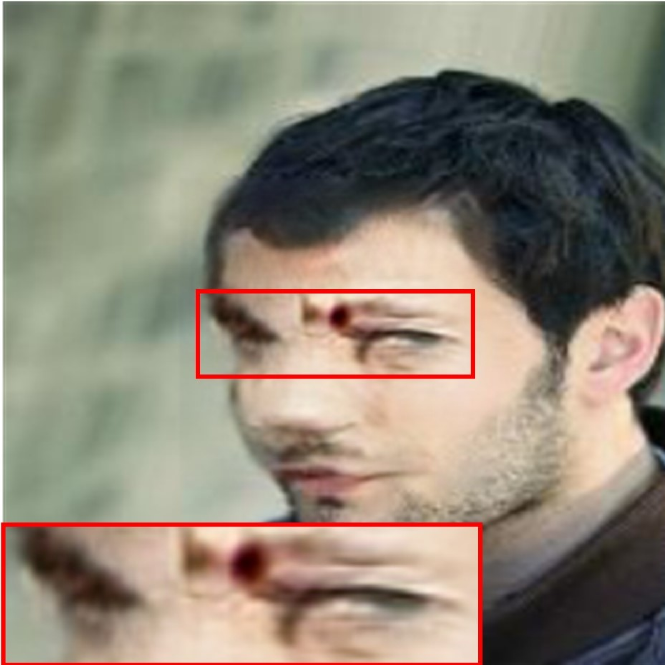}&
\includegraphics[width=\swtwo]{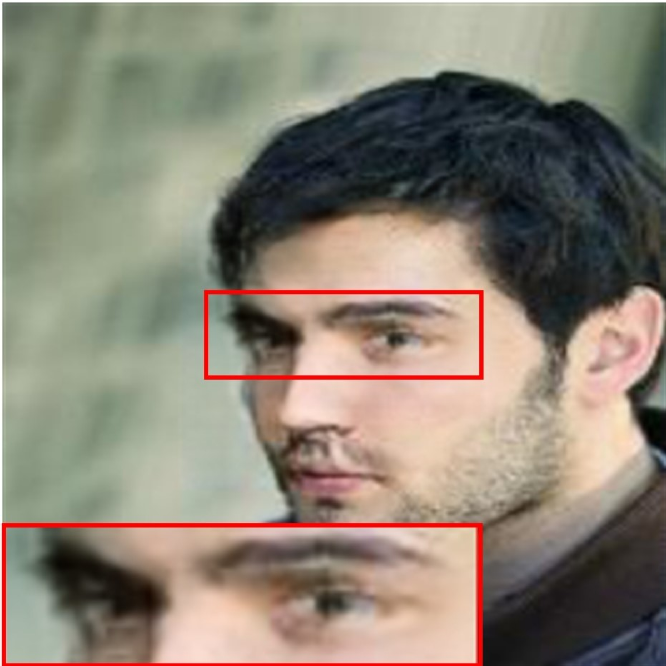}&
\includegraphics[width=\swtwo]{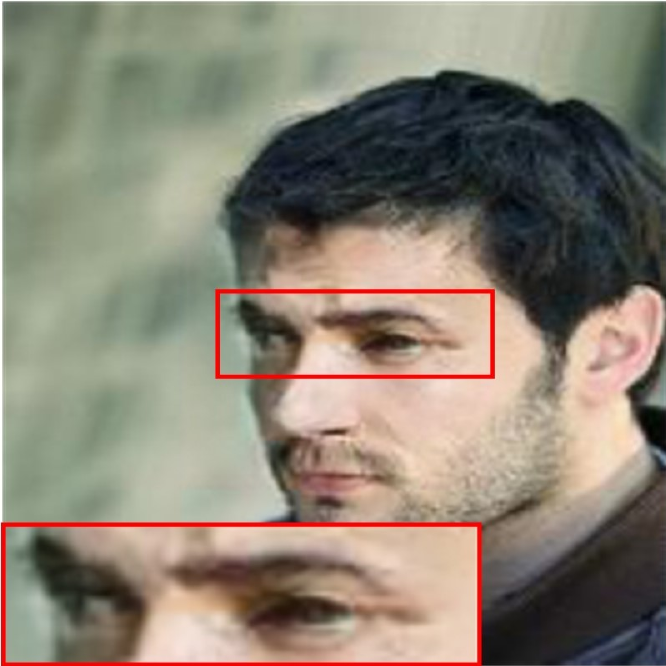}&
\includegraphics[width=\swtwo]{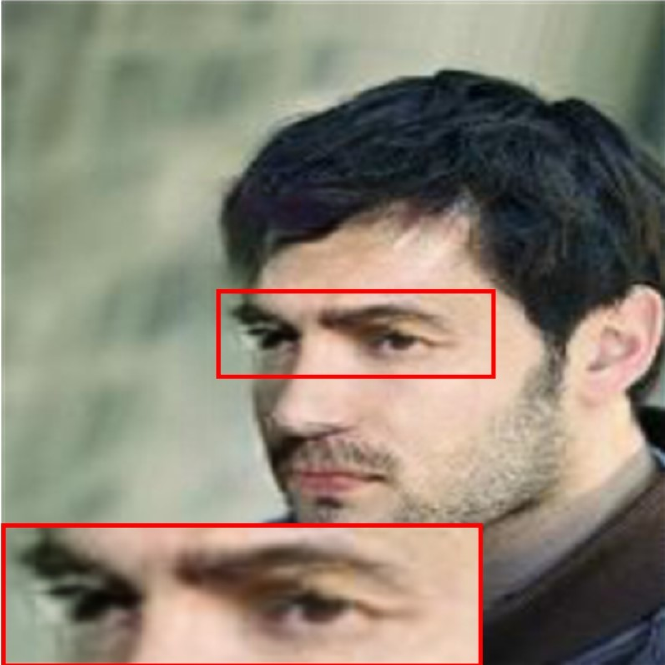}&
\includegraphics[width=\swtwo]{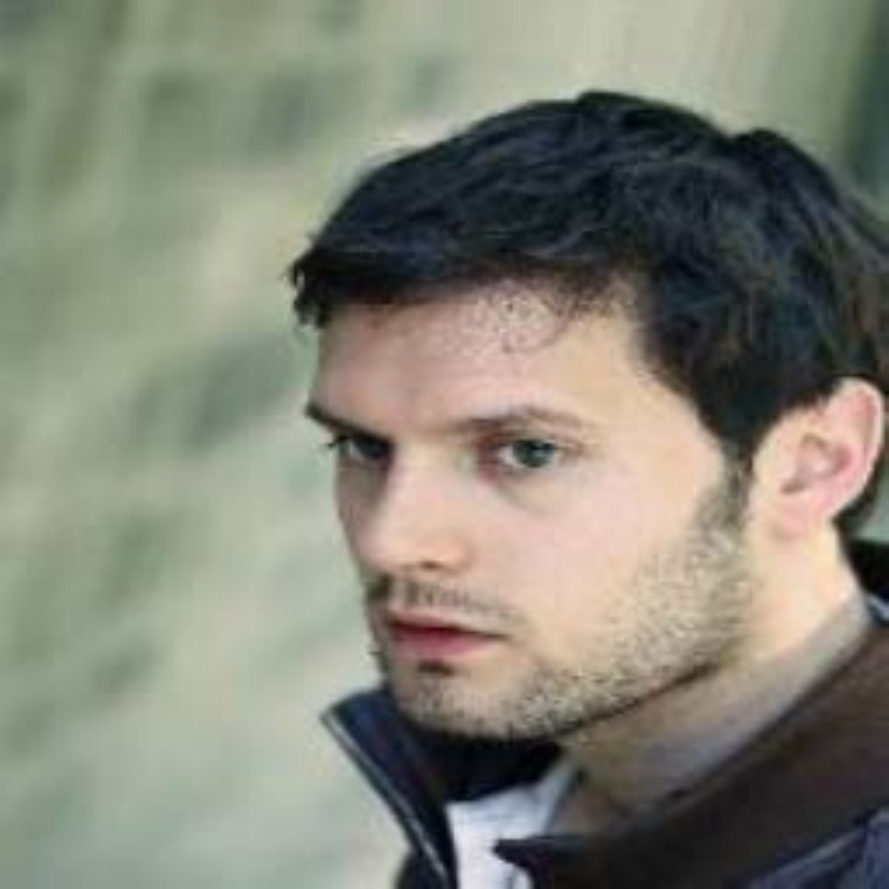}\\
(a) Input &(b) CE \cite{pathak2016context} &(c) CA~\cite{yu2018generative}&(d) SH~\cite{yan2018shift}&(e) MED~\cite{liu2020rethinking}&(f) Ours&(g) GT
\end{tabular}
\vspace{-2mm}
\caption{Visual comparison on the CelebA dataset.}
\label{imgcent}
\end{figure*}

\subsection{Discriminator} The discriminator consists of the global critic network and local critic network, which can ensure the local and global content consistency. Each critic network includes six convolution layers with the kernel size of 4 and stride of 2, and use the Leaky ReLu with the slope of 0.2 for all but the last layer. Furthermore, the spectral normalization is adopted in our network to achieve stable training.

\subsection{Loss Function}
Our network is trained with a series of loss functions, including pixel reconstruction loss, perceptual loss, style loss, and adversarial loss so that the finally generated image looks more visually realistic.

\textbf{Pixel Reconstruction Loss.} We adopt the $l_1$ distance as the reconstruction loss from two aspects. The first aspect is to supervise the structure and texture branches. The corresponding loss functions are formulated in Eqs. (\ref{con:structureBranch}) and (\ref{con:textureBranch}), respectively. The second aspect is to measure the similarity between the final output result $\boldsymbol{I}_{out}$ and the ground-truth image $\boldsymbol{I}_{gt}$ by
\begin{equation}
\footnotesize{
 \mathcal{L}_{\textrm{rec}}=\left\|\boldsymbol{I}_{out}-\boldsymbol{I}_{gt}\right\|_1.
}
\end{equation}

\textbf{Perceptual Loss.} We utilize the perceptual loss $\mathcal{L}_{\textrm{perc}}$ to capture the high-level semantics \cite{johnson2016perceptual} by computing the $l_1$ distance between the feature spaces of $\boldsymbol{I}_{out}$ and $\boldsymbol{I}_{gt}$  through ImageNet-pretrained VGG-16 backbone, which can be written as
\begin{equation}
\footnotesize{
 \mathcal{L}_{\textrm{perc}}=\mathbb{E}\left[\sum_{i}\left\|\phi_{i}\left(\boldsymbol{I}_{out}\right)-\phi_{i}\left(\boldsymbol{I}_{gt}\right)\right\|_1\right],
}
\end{equation}
where $\phi_{i}(\cdot), i=1,...,5$ denote the five activation maps from VGG-16, which are ReLu1\_1, ReLu2\_1, ReLu3\_1, ReLu4\_1 and ReLu5\_1.

\textbf{Style Loss.} We introduce the style loss $\mathcal{L}_{\textrm{style}}$ to mitigate style differences, which is defined as:
\begin{equation}
\footnotesize{
 \mathcal{L}_{\textrm{style}}=\mathbb{E}\left[\sum_{i}\left\|\varphi_{i}\left(\boldsymbol{I}_{out}\right)-\varphi_{i}\left(\boldsymbol{I}_{gt}\right)\right\|_1\right],
}
\end{equation}
where $\varphi_{i}(\cdot) = \phi_{i}^\mathrm{T}\phi_{i}$ denotes the Gram matrix constructed from the above-mentioned five activation maps.

\textbf{Adversarial Loss.} The relativistic average least squares adversarial loss $\mathcal{L}_{\textrm{adv}}$ is to ensure the local and global content consistency. For the generator, the adversarial loss is defined as:
\begin{equation}
\footnotesize{
 \mathcal{L}_{\textrm{adv}}=-\mathbb{E}_{\mathbf{x}_{r}}\left[\textrm{log}\left(1-D_{ra}\left(\mathbf{x}_{r},\mathbf{x}_{f}\right)\right)\right]-\mathbb{E}_{\mathbf{x}_{f}}\left[\textrm{log}\left(D_{ra}\left(\mathbf{x}_{f},\mathbf{x}_{r}\right)\right)\right],
}
\end{equation}
where $D_{ra}\left(\mathbf{x}_{r},\mathbf{x}_{f}\right) = \textrm{Sigmoid}\left(D_{gl}\left(\mathbf{x}_{r}\right)-\mathbb{E}_{\mathbf{x}_{f}}\left[D_{gl}\left(\mathbf{x}_{f}\right)\right]\right)$, $D_{gl}$ denotes the local or global discriminator without the last Sigmoid function, and $\left(\mathbf{x}_{r}, \mathbf{x}_{f}\right)$ denotes a pair of the ground-truth and output images.

\textbf{Total Loss.} The total loss of the proposed method can be obtained by
\begin{equation}
\footnotesize{
 \mathcal{L}_{\textrm{total}}=\lambda_{r}\mathcal{L}_{\textrm{rec}}+\lambda_{p}\mathcal{L}_{\textrm{prec}}+\lambda_{s}\mathcal{L}_{\textrm{style}}+\lambda_{adv}\mathcal{L}_{\textrm{adv}}+\lambda_{te}\mathcal{L}_{\textrm{rte}}+\lambda_{st}\mathcal{L}_{\textrm{rst}},
}
\end{equation}
where $\lambda_{r}$, $\lambda_{p}$, $\lambda_{s}$, $\lambda_{adv}$, $\lambda_{te}$ and $\lambda_{st}$ are the tradeoff parameters, and we empirically set $\lambda_{r} = 1$, $\lambda_{p} = 0.2$, $\lambda_{s} = 250$, $\lambda_{adv} = 0.2$, $\lambda_{te} = 1$ and $\lambda_{st} = 1$.

\section{Experimental Results}
\vspace{-1mm}
In our experiment, three public datasets are used for the verification, including Places2 \cite{zhou2017places},  Paris StreetView \cite{doersch2012makes}, and CelebA \cite{liu2015deep}. We follow the training, testing, and validation splits of these three datasets themselves. The irregular masks are taken from PConv \cite{liu2018image} and classified based on the ratios of the hole to the entire image with an increment of 10\%. Our network is built on the PyTorch framework, trained on a single NVIDIA 2080 Ti GPU (11GB), and optimized by the Adam optimizer with a learning rate of $2 \times 10^{-4}$. The CelebA, Paris StreetView, and Places2 models require around 10, 55, and 100 epochs, respectively. All the masks and images are resized to $256 \times 256$.

\begin{table*}[t!]
\setlength{\abovecaptionskip}{0mm}
\renewcommand\arraystretch{1.2}
\footnotesize
\caption{Performance comparison on the Places2 dataset.}
\vspace{-2mm}
\begin{center}
\setlength{\tabcolsep}{1mm}{
\scalebox{0.94}{
\begin{tabular}{|c|cccc|cccc|cccc|cccc|}
\hline
  Metrics &\multicolumn{4}{c|}{$\textrm{PSNR}^\mathrm{\uparrow}$}&\multicolumn{4}{c|}{$\textrm{SSIM}^\mathrm{\uparrow}$}&\multicolumn{4}{c|}{$\textrm{MAE}^\mathrm{\downarrow}$}&\multicolumn{4}{c|}{$\textrm{FID}^\mathrm{\downarrow}$}
  \\
  \hline
  Mask Ratio & 10-20\% & 20-30\% & 30-40\% & 40-50\% & 10-20\% & 20-30\% & 30-40\% & 40-50\% & 10-20\% & 20-30\% & 30-40\% & 40-50\% & 10-20\% & 20-30\% & 30-40\% & 40-50\%  \\
  \hline
  GC \cite{yu2019free} &29.16&26.35&24.24&22.35&0.928&0.880&0.791&0.704&0.0116&0.0195&0.0283&0.0391&28.31&39.01&49.81&62.73\\
  \hline
  EC \cite{nazeri2019edgeconnect} &30.44&27.46&25.69&24.02&0.939&0.886&0.830&0.759&0.0109&0.0186&0.0259&0.0350&18.41&31.00&42.03&54.85\\
  \hline
  MED \cite{liu2020rethinking} &30.88&27.54&25.40&23.55&0.945&0.890&0.828&0.750&0.0101&0.0180&0.0262&0.0365&20.18&36.68&50.14&65.72\\
  \hline
  RFR \cite{li2020recurrent} &31.04&27.77&25.77&24.09&0.945&0.893&0.836&0.768&0.0091&0.0166&0.0241&0.0330&16.22&29.32&41.05&\bf{53.22}\\
  \hline
  DSI \cite{peng2021generating} &31.02&27.59&25.45&23.61&0.945&0.890&0.829&0.752&0.0100&0.0179&0.0260&0.0359&\bf{15.79}&\bf{28.94}&\bf{40.87}&54.73 \\
  \hline
  Ours  &\bf{32.24}&\bf{28.68}&\bf{26.23}&\bf{24.32}&\bf{0.957}&\bf{0.912}&\bf{0.854} &\bf{0.783} &\bf{0.0086} &\bf{0.0153} &\bf{0.0233} &\bf{0.0326} &16.13&29.14&42.23 &58.01\\
  \hline
\end{tabular}
}
}
\end{center}
\label{tab:Places2_comparison}
\end{table*}

\begin{table*}[t!]
\setlength{\abovecaptionskip}{0mm}
\renewcommand\arraystretch{1.2}
\footnotesize
\caption{Performance comparison on the Paris StreetView dataset.}
\vspace{-2mm}
\begin{center}
\setlength{\tabcolsep}{1mm}{
\scalebox{0.94}{
\begin{tabular}{|c|cccc|cccc|cccc|cccc|}
\hline
  Metrics &\multicolumn{4}{c|}{$\textrm{PSNR}^\mathrm{\uparrow}$}&\multicolumn{4}{c|}{$\textrm{SSIM}^\mathrm{\uparrow}$}&\multicolumn{4}{c|}{$\textrm{MAE}^\mathrm{\downarrow}$}&\multicolumn{4}{c|}{$\textrm{FID}^\mathrm{\downarrow}$}
  \\
  \hline
  Mask Ratio & 10-20\% & 20-30\% & 30-40\% & 40-50\% & 10-20\% & 20-30\% & 30-40\% & 40-50\% & 10-20\% & 20-30\% & 30-40\% & 40-50\% & 10-20\% & 20-30\% & 30-40\% & 40-50\%  \\
  \hline
  GC \cite{yu2019free} &30.77&27.71&25.58&23.81&0.943&0.897&0.839&0.782&0.0158&0.0234&0.0313&0.0409&28.44&39.91&52.72&69.26\\
  \hline
  EC \cite{nazeri2019edgeconnect} &31.08&28.08&25.95&24.30&0.949&0.906&0.848&0.789&0.0110&0.0194&0.0283&0.0390&22.22&38.69&55.85&72.43\\
  \hline
  MED \cite{liu2020rethinking} &32.28&28.75&26.08&24.32&0.966&0.925&0.877&0.813&0.0105&0.0190&0.0282&0.0387&20.29&30.79&48.91&67.42\\
  \hline
  RFR \cite{li2020recurrent} &31.63&28.61&26.63&24.89&0.957&0.920&0.872&0.825&0.0152&0.0220&0.0296&0.0388&18.07&30.81&\bf{42.61}&\bf{51.80}\\
  \hline
  DSI \cite{peng2021generating} &31.42 &28.21 &26.04 &24.07 &0.953 &0.912 &0.866 &0.799 &0.0121 &0.0197 &0.0285 &0.0393 &18.09 &33.36 &47.80 &60.15 \\
  \hline
  Ours  &\bf{33.68}&\bf{30.29}&\bf{27.34}&\bf{25.28}&\bf{0.971}&\bf{0.940}&\bf{0.889} &\bf{0.828} &\bf{0.0090} &\bf{0.0152} &\bf{0.0240} &\bf{0.0336} &\bf{13.83} &\bf{27.32}&45.44 &56.07\\
  \hline
\end{tabular}
}
}
\end{center}
\label{tab:ParisStreetView_comparison}
\end{table*}

\begin{table}[t!]
\setlength{\abovecaptionskip}{0mm}
\renewcommand\arraystretch{1.2}
\footnotesize
\caption{Performance comparison on the CelebA dataset.}
\vspace{-2mm}
\begin{center}
\scalebox{0.94}{
\begin{tabular}{|c|c|c|c|c|}
  \hline
  Metrics &$\textrm{PSNR}^\mathrm{\uparrow}$ &$\textrm{SSIM}^\mathrm{\uparrow}$ &$\textrm{MAE}^\mathrm{\downarrow}$ &$\textrm{FID}^\mathrm{\downarrow}$\\
  \hline
  CE \cite{pathak2016context} &25.12 &0.899 &0.0352 &3.84\\
  \hline
  CA \cite{yu2018generative} &24.54 &0.887 &0.0286 &5.01 \\
  \hline
  SH \cite{yan2018shift} &26.03 &0.901 &0.0245 &1.71 \\
  \hline
  MED \cite{liu2020rethinking} &25.92 &0.914 &0.0241 &1.59 \\
  \hline
  Ours  &\bf{26.38}&\bf{0.922}&\bf{0.0223}&\bf{1.40}\\
  \hline
\end{tabular}
}
\end{center}
\label{tab:CelebA_comparison}
\end{table}

\subsection{Performance comparison with state-of-the-art}
We compare the proposed method with eight state-of-the-art methods, including GC \cite{yu2019free}, EC \cite{nazeri2019edgeconnect}, MED \cite{liu2020rethinking}, RFR \cite{li2020recurrent}, DSI \cite{peng2021generating}, CE \cite{pathak2016context}, CA \cite{yu2018generative} and SH \cite{yan2018shift}. For the evaluation metrics, we use four common metrics: PSNR, SSIM, MAE and FID (Fr\'{e}chet Inception Distance) \cite{heusel2017gans}. For fair comparison, we use the same experimental setup for all the compared methods. The experiments are conducted on two types of damaged images containing center hole and irregular hole. In the following, we present the experimental results and analysis in detail.

For images with irregular holes, we take the Places2 \cite{zhou2017places} and Paris StreetView \cite{doersch2012makes} datasets for evaluation, and select several representative methods GC \cite{yu2019free}, EC \cite{nazeri2019edgeconnect}, MED \cite{liu2020rethinking}, RFR \cite{li2020recurrent} and DSI \cite{peng2021generating} for performance comparison. Meanwhile, we use the same validation images as the MED method. Experimental results are shown in Tables \ref{tab:Places2_comparison} and \ref{tab:ParisStreetView_comparison}. We can see from the two tables that our method achieves significant improvement over the compared methods in terms of the PSNR, SSIM, and MAE metrics. This is due to the fact that our method designs two effective multi-feature fusion techniques, which not only exploits the connection between the structure and texture features, but also considers the relationship within image context. For the FID metric, our method still shows competitive performance on the images with small damaged region. When evaluated on the images with large damaged region, our method exhibits minor performance degradation. The reason might be as follows. The FID metric aims to measure the distance between the feature spaces of two groups of images. While the proposed method does not introduce the structure prior in the encoding stage, thus enlarging the distance between the feature spaces of the generated output and ground-true images. Besides, we can observe from Fig. \ref{imgirre} that the images inpainted by the proposed method have better visual quality than those of all the compared methods.

For images with center hole, we compare the proposed method with four typical methods including CE \cite{pathak2016context}, CA \cite{yu2018generative}, SH \cite{yan2018shift} and MED \cite{liu2020rethinking}. The performance is evaluated on 10,000 images selected randomly from the CelebA \cite{liu2015deep} validation dataset. The result is shown in Table \ref{tab:CelebA_comparison}. We can see that the proposed method still performs best. Especially for the FID metric, our method can better infer the missing structure and texture of an image compared with its competitors.
The subjective quality of the inpainted images is shown in Fig. \ref{imgcent}. It can be observed that our method demonstrates its effectiveness in dealing with the center hole images.

\begin{table}[t!]
\setlength{\abovecaptionskip}{0mm}
\renewcommand\arraystretch{1.2}
\footnotesize
\caption{Ablation study of different modules on Paris StreetView. Here, random masks with mask ratio 30\%-40\% are used.}
\vspace{-2mm}
\begin{center}
\scalebox{0.94}{
\begin{tabular}{|c|c|c|c|c|}
  \hline
  Metrics &$\textrm{PSNR}^\mathrm{\uparrow}$ &$\textrm{SSIM}^\mathrm{\uparrow}$ &$\textrm{MAE}^\mathrm{\downarrow}$ &$\textrm{FID}^\mathrm{\downarrow}$\\
  \hline
  w/o SDFF &27.15 &0.883 &0.0244 &46.33 \\
  \hline
  w/o CA &27.08 &0.879 &0.0249 &\bf{44.56} \\
  \hline
  w/o SKNet &27.17 &0.885 &0.0242 &45.61 \\
  \hline
  w/o BPFA &26.99 &0.880 &0.0247 &46.39 \\
  \hline
  w/o PC &26.93 &0.877 &0.0250 &46.92 \\
  \hline
  Baseline &26.08 &0.877 &0.0282 &48.91 \\
  \hline
  Ours  &\bf{27.34}&\bf{0.889}&\bf{0.0240}&45.44\\
  \hline
\end{tabular}
}
\end{center}
\label{tab:Ablation_comparison}
\end{table}

\subsection{Ablation study}
To verify the contribution of each individual component of our network, the ablation study is performed on the Paris StreetView dataset. The result is shown in Table \ref{tab:Ablation_comparison}. We can see that each module shows its effectiveness. Specifically speaking, the partial convolution based padding (PC) module \cite{liu2018image} improves the performance by making the model pay more attention to visible pixels (undamaged regions). Moreover, our designed BPFA module contributes to the whole network the best. This is because BPFA enhances the connection from local feature to overall consistency, thus reducing visible artifacts and unpleasant contents. In addition, we can find that the SDFF module has the second contribution. As expected, SDFF captures the correlation between the structure and texture features. Interestingly, the CA module can learn contextual feature representations and the SKNet module can effectively equalize the structure and texture features generated by the multi-scale filling stage, thereby benefiting the proposed network.

To better visulization, we show the outputs and ground-truths for Eqs. (1) and (2) in Fig. \ref{eqs1and2}. Fig. \ref{eqs1and2} (a) and (c) are our structure and texture feature maps obtained after the multi-scale filling stage, respectively. Fig. \ref{eqs1and2} (b) and (d) are the structure and texture of ground-turth, respectively. We can see that deep-layer convolution focuses more on image structure (Fig. \ref{eqs1and2} (a) and (b)) and shallow-layer convolution focuses more on image texture (Fig. \ref{eqs1and2} (c) and (d)).

\def\swtwo{0.35\linewidth}
\begin{figure}[h!]
\centering
\begin{tabular}{cccc}
\includegraphics[width=\swtwo]{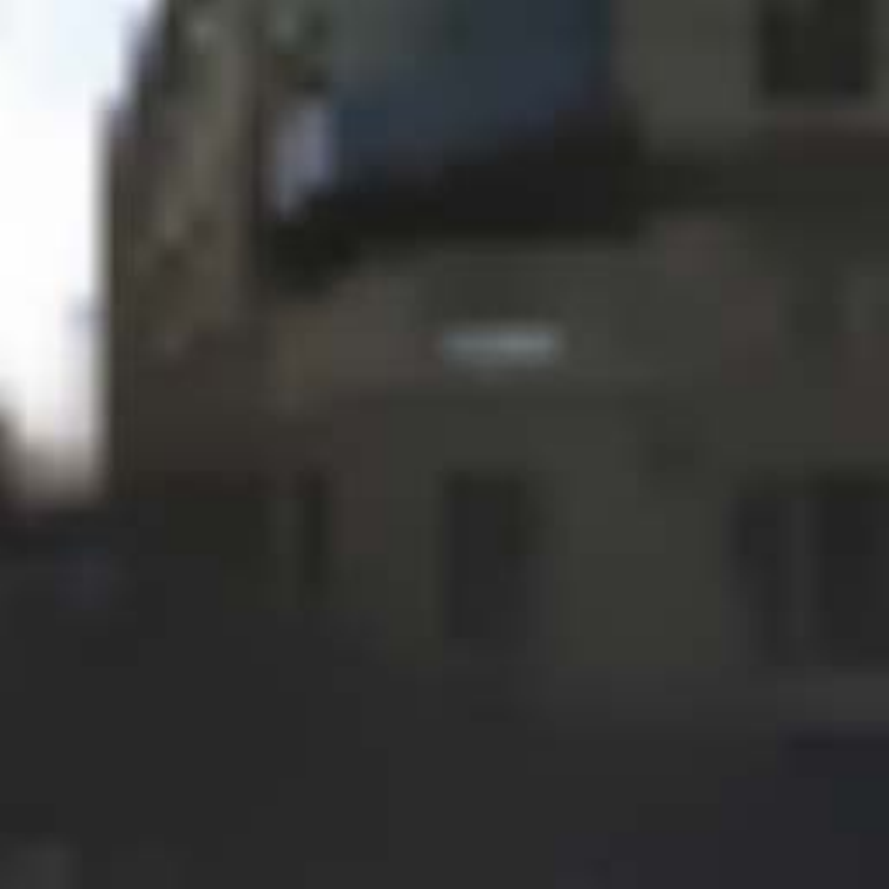}&
\includegraphics[width=\swtwo]{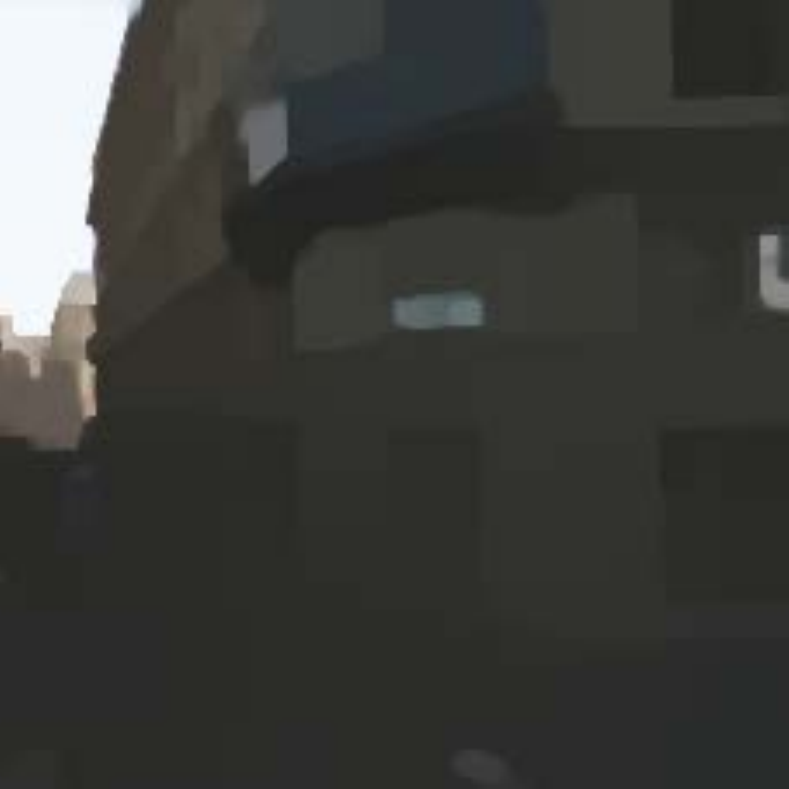}
\\
(a) Our structure & (b) GT structure & \\
\includegraphics[width=\swtwo]{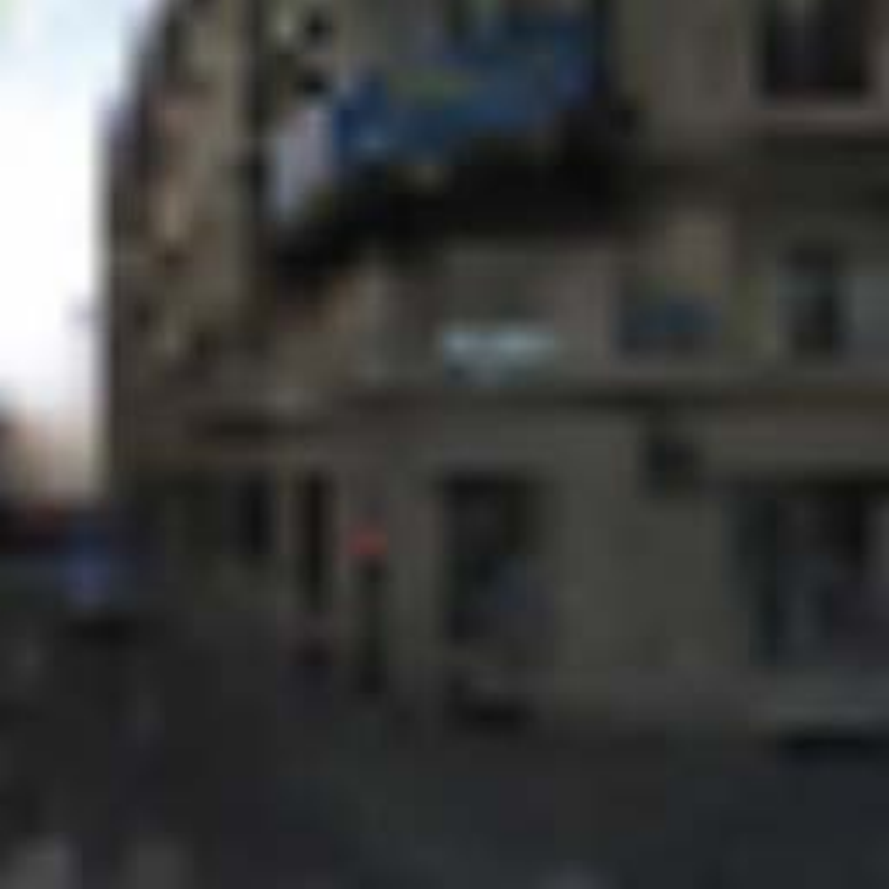}&
\includegraphics[width=\swtwo]{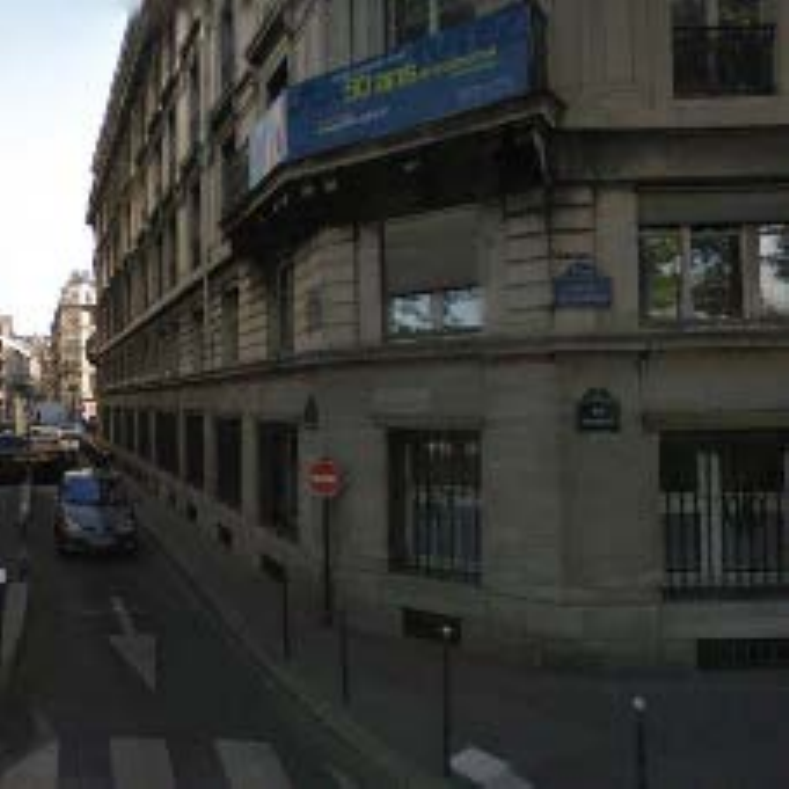}\\
(c) Our texture & (d) GT texture
\end{tabular}
\caption{Visual outputs for  Eqs. (1) and (2).}
\label{eqs1and2}
\end{figure}

In addition, according to our ablation study, our method with SDFF performs better than that without SDFF. This shows that our SDFF is a better channel attention module than existing channel attention modules. An example of the feature output is shown in Fig. \ref{imgGate}. It can be observed Fig. \ref{imgGate} that these two soft gates can respectively response the structure and texture well. It is worth emphasizing that our method use $\mathcal{L}_{\textrm{rst}}$ and $\mathcal{L}_{\textrm{rte}}$ to make deep-layer and shallow-layer convolutions focus on structure and texture features, respectively. Meanwhile, $\mathcal{L}_{\textrm{perc}}$ and $\mathcal{L}_{\textrm{style}}$ are used to improve perceptual quality and mitigate style differences, respectively. Clearly, $\mathcal{L}_{\textrm{rec}}$ and $\mathcal{L}_{\textrm{adv}}$ are necessary for pixel reconstruction and adversarial training. The ablation experiments on these losses are shown in Table \ref{tab:AblationLoss_comparison}. It is clear that each loss contributes to our method.

\def\swtwo{0.325\linewidth}
\begin{figure}[h!]
\renewcommand{\tabcolsep}{1pt}
\begin{tabular}{ccccccc}
\includegraphics[width=\swtwo]{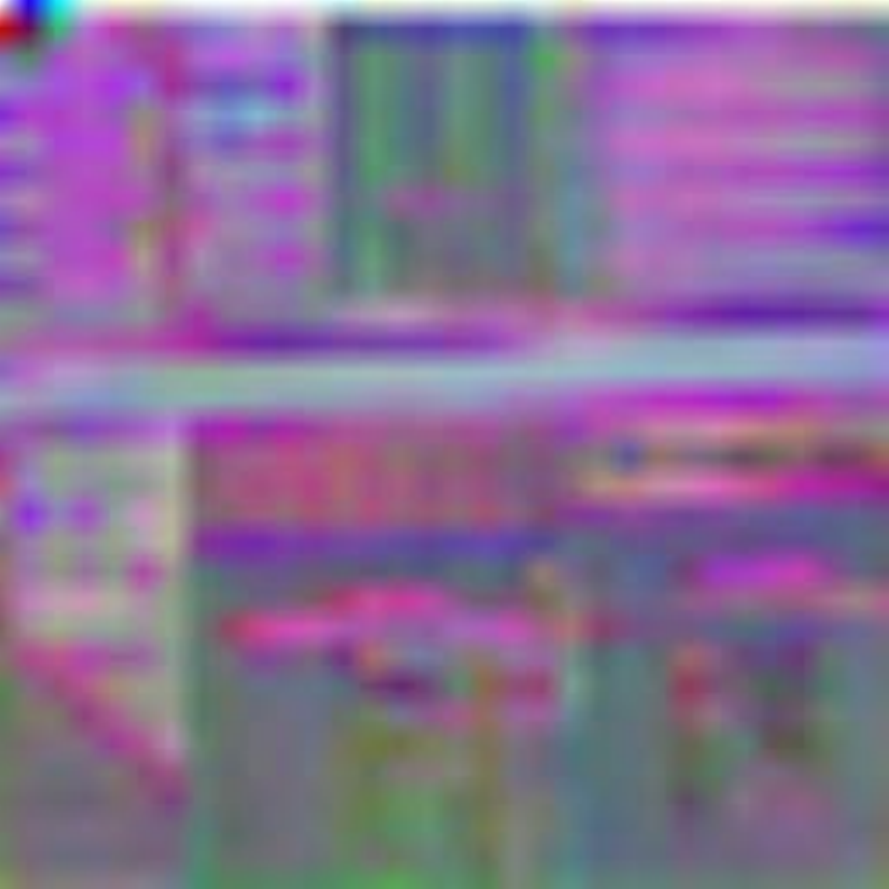}&
\includegraphics[width=\swtwo]{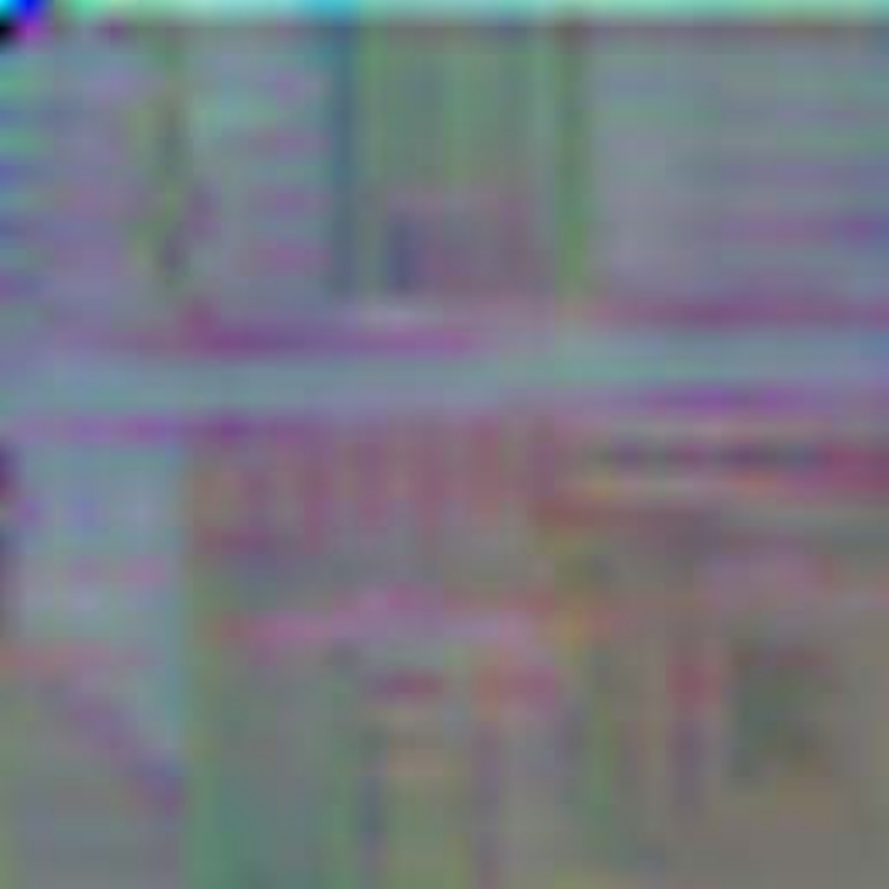}&
\includegraphics[width=\swtwo]{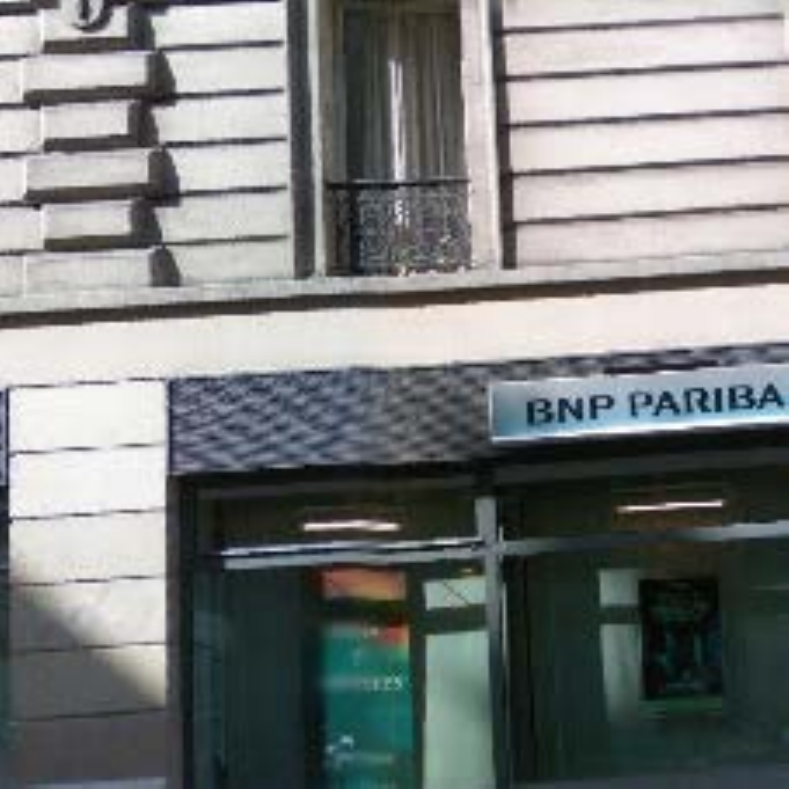}\\
(a) Structure & (b) Texture & (c) Ground Truth
\end{tabular}
\caption{Visualization of the feature maps for the structure and texture information.}
\label{imgGate}
\end{figure}

\begin{table}[h!]
\setlength{\abovecaptionskip}{0mm}
\renewcommand\arraystretch{1.2}
\footnotesize
\caption{Ablation study of different losses on Paris StreetView. Here, random masks with mask ratio 30\%-40\% are used.}
\vspace{-1mm}
\begin{center}
\scalebox{1}{
\begin{tabular}{|c|c|c|c|c|}
  \hline
  Metrics &$\textrm{PSNR}^\mathrm{\uparrow}$ &$\textrm{SSIM}^\mathrm{\uparrow}$ &$\textrm{MAE}^\mathrm{\downarrow}$ &$\textrm{FID}^\mathrm{\downarrow}$\\
  \hline
  w/o $\mathcal{L}_{\textrm{perc}}$ &26.99 &0.879 &0.0254 &46.96 \\
  \hline
  w/o $\mathcal{L}_{\textrm{style}}$ &26.98 &0.879 &0.0250 &47.55 \\
  \hline
  w/o $\mathcal{L}_{\textrm{rst}}$ &27.16 &0.883 &0.0251 &45.95 \\
\hline
  w/o $\mathcal{L}_{\textrm{rte}}$ &27.23 &0.887 &0.0252 &\bf{43.96} \\
  \hline
  Ours  &\bf{27.34}&\bf{0.889}&\bf{0.0240}&45.44\\
  \hline
\end{tabular}
}
\end{center}
\label{tab:AblationLoss_comparison}
\vspace{-4mm}
\end{table}


\section{Conclusion}
In this paper, we have presented a deep multi-feature co-learning network for image inpainting, which can yield the detailed textures and reasonable structures. Our network designs two new fusion modules:  Soft-gating Dual Feature Fusion (SDFF) and Bilateral Propagation Feature Aggregation (BPFA). SDFF can control the fusion ratio through a soft gating technique to refine the structure and texture features,  making the structures and textures more consistent. BPFA can co-learn contextual attention, channel-wise information and feature space. By the co-learning strategy, the inpainted images preserve the from local to overall consistency.

Note that the soft gating operation is commonly used to channel attention. But our SDFF module exploits two soft gates ($\boldsymbol{G}_{st}$ and $\boldsymbol{G}_{te}$) to make effective response to structure and texture information, respectively. Through $\boldsymbol{G}_{st}$ and $\boldsymbol{G}_{te}$, our SDFF can control the degree of integrating structure and texture information. We also tried to first concatenate $\boldsymbol{F}_{cst}$ and $\boldsymbol{F}_{cte}$, and then used channel attention module (SENet [8]) directly to output feature. However, this will lead to dynamic output changes and impair the training of GAN. Our experiment also shows that this strategy achieves only limited improvement for our model. Therefore, we propose to multiply reweighted $\boldsymbol{F}_{cte}$ by $\boldsymbol{F}_{cst}$ to selectively transmit useful features, resulting in significant performance improvement. In the future, we will explore the performance of introducing the structure prior in the encoding stage.


{\small
\nocite{*}
\bibliographystyle{ieee}
\bibliography{linbib.bib}

\begin{thebibliography}{10}\itemsep=-1pt

\bibitem{barnes2009patchmatch}
C.~Barnes, E.~Shechtman, A.~Finkelstein, and D.~B. Goldman.
\newblock Patchmatch: A randomized correspondence algorithm for structural
  image editing.
\newblock {\em ACM Transactions on Graphics (ToG)}, 28(3):1--11, 2009.

\bibitem{bertalmio2000image}
M.~Bertalmio, G.~Sapiro, V.~Caselles, and C.~Ballester.
\newblock Image inpainting.
\newblock In {\em Proceedings of the 27th Annual Conference on Computer
  Graphics and Interactive Techniques (SIGGRAPH)}, pages 417--424, 2000.

\bibitem{darabi2012image}
S.~Darabi, E.~Shechtman, C.~Barnes, D.~B. Goldman, and P.~Sen.
\newblock Image melding: Combining inconsistent images using patch-based
  synthesis.
\newblock {\em ACM Transactions on Graphics (ToG)}, 31(4):1--10, 2012.

\bibitem{doersch2012makes}
C.~Doersch, S.~Singh, A.~Gupta, J.~Sivic, and A.~Efros.
\newblock What makes paris look like paris?
\newblock {\em ACM Transactions on Graphics (ToG)}, 31(4):1--9, 2012.

\bibitem{efros2001image}
A.~A. Efros and W.~T. Freeman.
\newblock Image quilting for texture synthesis and transfer.
\newblock In {\em Proceedings of the 28th Annual Conference on Computer
  Graphics and Interactive Techniques (SIGGRAPH)}, pages 341--346, 2001.

\bibitem{hays2007scene}
J.~Hays and A.~A. Efros.
\newblock Scene completion using millions of photographs.
\newblock {\em ACM Transactions on Graphics (ToG)}, 26(3):1--7, 2007.

\bibitem{heusel2017gans}
M.~Heusel, H.~Ramsauer, T.~Unterthiner, B.~Nessler, and S.~Hochreiter.
\newblock Gans trained by a two time-scale update rule converge to a local nash
  equilibrium.
\newblock In {\em Proceedings of the Advances in Neural Information Processing
  Systems (NIPS)}, pages 6629--6640, 2017.

\bibitem{hu2018squeeze}
J.~Hu, L.~Shen, and G.~Sun.
\newblock Squeeze-and-excitation networks.
\newblock In {\em Proceedings of the IEEE Conference on Computer Vision and
  Pattern Recognition (CVPR)}, pages 7132--7141, 2018.

\bibitem{iizuka2017globally}
S.~Iizuka, E.~Simo-Serra, and H.~Ishikawa.
\newblock Globally and locally consistent image completion.
\newblock {\em ACM Transactions on Graphics (ToG)}, 36(4):1--14, 2017.

\bibitem{johnson2016perceptual}
J.~Johnson, A.~Alahi, and L.~Fei-Fei.
\newblock Perceptual losses for real-time style transfer and super-resolution.
\newblock In {\em Proceedings of the European Conference on Computer Vision
  (ECCV)}, pages 694--711, 2016.

\bibitem{lahiri2020prior}
A.~Lahiri, A.~K. Jain, S.~Agrawal, P.~Mitra, and P.~K. Biswas.
\newblock Prior guided gan based semantic inpainting.
\newblock In {\em Proceedings of the IEEE Conference on Computer Vision and
  Pattern Recognition (CVPR)}, pages 13696--13705, 2020.

\bibitem{li2021faceinpainter}
J.~Li, Z.~Li, J.~Cao, X.~Song, and R.~He.
\newblock Faceinpainter: High fidelity face adaptation to heterogeneous
  domains.
\newblock In {\em Proceedings of the IEEE Conference on Computer Vision and
  Pattern Recognition (CVPR)}, pages 5089--5098, 2021.

\bibitem{li2020recurrent}
J.~Li, N.~Wang, L.~Zhang, B.~Du, and D.~Tao.
\newblock Recurrent feature reasoning for image inpainting.
\newblock In {\em Proceedings of the IEEE Conference on Computer Vision and
  Pattern Recognition (CVPR)}, pages 7760--7768, 2020.

\bibitem{li2019selective}
X.~Li, W.~Wang, X.~Hu, and J.~Yang.
\newblock Selective kernel networks.
\newblock In {\em Proceedings of the IEEE Conference on Computer Vision and
  Pattern Recognition (CVPR)}, pages 510--519, 2019.

\bibitem{li2017generative}
Y.~Li, S.~Liu, J.~Yang, and M.-H. Yang.
\newblock Generative face completion.
\newblock In {\em Proceedings of the IEEE Conference on Computer Vision and
  Pattern Recognition (CVPR)}, pages 3911--3919, 2017.

\bibitem{liao2021image}
L.~Liao, J.~Xiao, Z.~Wang, C.-W. Lin, and S.~Satoh.
\newblock Image inpainting guided by coherence priors of semantics and
  textures.
\newblock In {\em Proceedings of the IEEE Conference on Computer Vision and
  Pattern Recognition (CVPR)}, pages 6539--6548, 2021.

\bibitem{liu2018image}
G.~Liu, F.~A. Reda, K.~J. Shih, T.-C. Wang, A.~Tao, and B.~Catanzaro.
\newblock Image inpainting for irregular holes using partial convolutions.
\newblock In {\em Proceedings of the European Conference on Computer Vision
  (ECCV)}, pages 85--100, 2018.

\bibitem{liu2020rethinking}
H.~Liu, B.~Jiang, Y.~Song, W.~Huang, and C.~Yang.
\newblock Rethinking image inpainting via a mutual encoder-decoder with feature
  equalizations.
\newblock In {\em Proceedings of the European Conference on Computer Vision
  (ECCV)}, pages 725--741, 2020.

\bibitem{liu2019coherent}
H.~Liu, B.~Jiang, Y.~Xiao, and C.~Yang.
\newblock Coherent semantic attention for image inpainting.
\newblock In {\em Proceedings of the IEEE International Conference on Computer
  Vision (ICCV)}, pages 4170--4179, 2019.

\bibitem{liu2021pd}
H.~Liu, Z.~Wan, W.~Huang, Y.~Song, X.~Han, and J.~Liao.
\newblock Pd-gan: Probabilistic diverse gan for image inpainting.
\newblock In {\em Proceedings of the IEEE Conference on Computer Vision and
  Pattern Recognition (CVPR)}, pages 9371--9381, 2021.

\bibitem{liu2015deep}
Z.~Liu, P.~Luo, X.~Wang, and X.~Tang.
\newblock Deep learning face attributes in the wild.
\newblock In {\em Proceedings of the IEEE International Conference on Computer
  Vision (ICCV)}, pages 3730--3738, 2015.

\bibitem{nazeri2019edgeconnect}
K.~Nazeri, E.~Ng, T.~Joseph, F.~Z. Qureshi, and M.~Ebrahimi.
\newblock Edgeconnect: Generative image inpainting with adversarial edge
  learning.
\newblock In {\em Proceedings of the IEEE International Conference on Computer
  Vision Workshop (ICCVW)}, pages 3265--3274, 2019.

\bibitem{pathak2016context}
D.~Pathak, P.~Krahenbuhl, J.~Donahue, T.~Darrell, and A.~A. Efros.
\newblock Context encoders: Feature learning by inpainting.
\newblock In {\em Proceedings of the IEEE Conference on Computer Vision and
  Pattern Recognition (CVPR)}, pages 2536--2544, 2016.

\bibitem{peng2021generating}
J.~Peng, D.~Liu, S.~Xu, and H.~Li.
\newblock Generating diverse structure for image inpainting with hierarchical
  vq-vae.
\newblock In {\em Proceedings of the IEEE Conference on Computer Vision and
  Pattern Recognition (CVPR)}, pages 10775--10784, 2021.

\bibitem{ren2019structureflow}
Y.~Ren, X.~Yu, R.~Zhang, T.~H. Li, S.~Liu, and G.~Li.
\newblock Structureflow: Image inpainting via structure-aware appearance flow.
\newblock In {\em Proceedings of the IEEE International Conference on Computer
  Vision (ICCV)}, pages 181--190, 2019.

\bibitem{tomasi1998bilateral}
C.~Tomasi and R.~Manduchi.
\newblock Bilateral filtering for gray and color images.
\newblock In {\em Proceedings of the IEEE International Conference on Computer
  Vision (ICCV)}, pages 839--846, 1998.

\bibitem{wang2021image}
T.~Wang, H.~Ouyang, and Q.~Chen.
\newblock Image inpainting with external-internal learning and monochromic
  bottleneck.
\newblock In {\em Proceedings of the IEEE Conference on Computer Vision and
  Pattern Recognition (CVPR)}, pages 5120--5129, 2021.

\bibitem{wang2018non}
X.~Wang, R.~Girshick, A.~Gupta, and K.~He.
\newblock Non-local neural networks.
\newblock In {\em Proceedings of the IEEE Conference on Computer Vision and
  Pattern Recognition (CVPR)}, pages 7794--7803, 2018.

\bibitem{xie2019image}
C.~Xie, S.~Liu, C.~Li, M.-M. Cheng, W.~Zuo, X.~Liu, S.~Wen, and E.~Ding.
\newblock Image inpainting with learnable bidirectional attention maps.
\newblock In {\em Proceedings of the IEEE International Conference on Computer
  Vision (ICCV)}, pages 8858--8867, 2019.

\bibitem{xiong2019foreground}
W.~Xiong, J.~Yu, Z.~Lin, J.~Yang, X.~Lu, C.~Barnes, and J.~Luo.
\newblock Foreground-aware image inpainting.
\newblock In {\em Proceedings of the IEEE Conference on Computer Vision and
  Pattern Recognition (CVPR)}, pages 5840--5848, 2019.

\bibitem{xu2012structure}
L.~Xu, Q.~Yan, Y.~Xia, and J.~Jia.
\newblock Structure extraction from texture via relative total variation.
\newblock {\em ACM Transactions on Graphics (ToG)}, 31(6):1--10, 2012.

\bibitem{yan2018shift}
Z.~Yan, X.~Li, M.~Li, W.~Zuo, and S.~Shan.
\newblock Shift-net: Image inpainting via deep feature rearrangement.
\newblock In {\em Proceedings of the European Conference on Computer Vision
  (ECCV)}, pages 1--17, 2018.

\bibitem{yang2020learning}
J.~Yang, Z.~Qi, and Y.~Shi.
\newblock Learning to incorporate structure knowledge for image inpainting.
\newblock In {\em Proceedings of the AAAI Conference on Artificial Intelligence
  (AAAI)}, pages 12605--12612, 2020.

\bibitem{yu2018generative}
J.~Yu, Z.~Lin, J.~Yang, X.~Shen, X.~Lu, and T.~S. Huang.
\newblock Generative image inpainting with contextual attention.
\newblock In {\em Proceedings of the IEEE Conference on Computer Vision and
  Pattern Recognition (CVPR)}, pages 5505--5514, 2018.

\bibitem{yu2019free}
J.~Yu, Z.~Lin, J.~Yang, X.~Shen, X.~Lu, and T.~S. Huang.
\newblock Free-form image inpainting with gated convolution.
\newblock In {\em Proceedings of the IEEE International Conference on Computer
  Vision (ICCV)}, pages 4471--4480, 2019.

\bibitem{yu2020region}
T.~Yu, Z.~Guo, X.~Jin, S.~Wu, Z.~Chen, W.~Li, Z.~Zhang, and S.~Liu.
\newblock Region normalization for image inpainting.
\newblock In {\em Proceedings of the AAAI Conference on Artificial Intelligence
  (AAAI)}, pages 12733--12740, 2020.

\bibitem{zeng2020high}
Y.~Zeng, Z.~Lin, J.~Yang, J.~Zhang, E.~Shechtman, and H.~Lu.
\newblock High-resolution image inpainting with iterative confidence feedback
  and guided upsampling.
\newblock In {\em Proceedings of the European Conference on Computer Vision
  (ECCV)}, pages 1--17, 2020.

\bibitem{zhou2017places}
B.~Zhou, A.~Lapedriza, A.~Khosla, A.~Oliva, and A.~Torralba.
\newblock Places: A 10 million image database for scene recognition.
\newblock {\em IEEE Transactions on Pattern Analysis and Machine Intelligence
  (TPAMI)}, 40(6):1452--1464, 2017.

\bibitem{zhou2021transfill}
Y.~Zhou, C.~Barnes, E.~Shechtman, and S.~Amirghodsi.
\newblock Transfill: Reference-guided image inpainting by merging multiple
  color and spatial transformations.
\newblock In {\em Proceedings of the IEEE Conference on Computer Vision and
  Pattern Recognition (CVPR)}, pages 2266--2276, 2021.

\end{thebibliography}
}



\end{document}